\documentclass[10pt,twocolumn,letterpaper]{article}

\usepackage[pagenumbers]{cvpr} 

\usepackage{pifont}  
\usepackage{xcolor}  

\definecolor{cvprblue}{rgb}{0.21,0.49,0.74}
\usepackage[pagebackref,breaklinks,colorlinks,allcolors=cvprblue]{hyperref}

\usepackage{amsmath}
\usepackage{times}
\usepackage{latexsym}
\usepackage{float}
\usepackage[utf8]{inputenc} 
\usepackage[T1]{fontenc}    
\usepackage{url}            
\usepackage{booktabs}       
\usepackage{wrapfig}
\usepackage{amsfonts}       
\usepackage{nicefrac}       
\usepackage{microtype}      
\usepackage{xcolor}         
\usepackage{multirow}
\usepackage{graphicx}
\usepackage{makecell} 
\usepackage{booktabs}
\usepackage{multirow}
\usepackage{color, colortbl}
\usepackage{subcaption}
\usepackage{tabu}
\usepackage{pifont}
\usepackage{makecell}
\usepackage{threeparttable}
\usepackage{enumitem}
\usepackage{hhline}
\usepackage{scalerel,xparse}
\usepackage{ulem}
\usepackage{caption}
\usepackage{pgfplots}
\usepackage{diagbox}
\usepackage{array}
\usepackage{color, colortbl}
\definecolor{Gray}{gray}{0.5}
\definecolor{LGray}{gray}{0.9}
\definecolor{darkblue}{RGB}{94,110,186}
\definecolor{darkGreen}{RGB}{92, 148, 110}
\definecolor{myblue}{RGB}{14, 121, 178}
\definecolor{myred}{RGB}{192, 0, 0}

\newcommand{\gray}[1]{\textcolor{gray}{#1}}

\newcommand{\darkGreen}[1]{\textcolor{darkGreen}{#1}}

\usepackage{adjustbox}
\usepackage{amssymb}

\definecolor{darkergreen}{RGB}{0,100,0}
\definecolor{myred}{rgb}{1,0,0}
\definecolor{myblue}{rgb}{0,0,1}
\definecolor{mygreen}{rgb}{0,1,0}
\definecolor{myyellow}{rgb}{1,1,0}

\definecolor{my_green}{RGB}{51,102,0}
\definecolor{my_red}{RGB}{204, 0, 0}
\renewcommand{\checkmark}{\textcolor{my_green}{\ding{51}}} 
\newcommand{\crossmark}{\textcolor{my_red}{\ding{55}}} 

\usepackage[pagebackref,breaklinks,colorlinks]{hyperref}

\usepackage[capitalize]{cleveref}
\crefname{section}{Sec.}{Secs.}
\Crefname{section}{Section}{Sections}
\Crefname{table}{Table}{Tables}
\crefname{table}{Tab.}{Tabs.}

\title{MLVU: Benchmarking Multi-task Long Video Understanding}

\author{
 Junjie Zhou\thanks{Co-first authors}~~$^{1,2}$, Yan Shu$^{*}$$^{1}$, Bo Zhao$^{*}$$^{1,3}$, Boya Wu$^{1}$, Zhengyang Liang$^{1}$, Shitao Xiao$^{1}$,\\
 Minghao Qin$^{1}$, Xi Yang$^{1}$, Yongping Xiong$^{2}$, Bo Zhang$^{4}$, Tiejun Huang$^{1,5}$, Zheng Liu\thanks{Corresponding author}~~$^{1}$ \\
 $^{1}$ Beijing Academy of Artificial Intelligence, $^{2}$ Beijing University of Posts and Telecommunications, \\
 $^{3}$ Shanghai Jiao Tong University, $^{4}$ Zhejiang University, $^{5}$ Peking University \\ 
 \texttt{\{junjiebupt, bozhaonanjing, zhengliu1026\}@gmail.com}
}

\begin{document}
\maketitle
\begin{abstract}
   The evaluation of Long Video Understanding (LVU) performance poses an important but challenging research problem. Despite previous efforts, the existing video understanding benchmarks are severely constrained by several issues, especially the insufficient lengths of videos, a lack of diversity in video types and evaluation tasks, and the inappropriateness for evaluating LVU performances. To address the above problems, we propose a new benchmark called MLVU (Multi-task Long Video Understanding Benchmark) for the comprehensive and in-depth evaluation of LVU. MLVU presents the following critical values: \textit{1)} The substantial and flexible extension of video lengths, which enables the benchmark to evaluate LVU performance across a wide range of durations. \textit{2)} The inclusion of various video genres, e.g., movies, surveillance footage, egocentric videos, cartoons, game videos, etc., which reflects the models' LVU performances in different scenarios. \textit{3)} The development of diversified evaluation tasks, which enables a comprehensive examination of MLLMs' key abilities in long-video understanding. The empirical study with 23 latest MLLMs reveals significant room for improvement in today's technique, as all existing methods struggle with most of the evaluation tasks and exhibit severe performance degradation when handling longer videos. Additionally, it suggests that factors such as context length, image-understanding ability, and the choice of LLM backbone can play critical roles in future advancements. We anticipate that MLVU will advance the research of long video understanding by providing a comprehensive and in-depth analysis of MLLMs. 

\end{abstract}
\vspace{-0.3cm}    
\vspace{-0.3cm}
\section{Introduction}
\label{intro}

Large language models (LLMs) are growing into a general solution for numerous AI tasks \cite{brown2020language,touvron2023llama}. In recent years, it becomes increasingly emphasized to extend LLMs with multi-modal capabilities and thus bring the Multi-modal LLM, namely, MLLM. Remarkably, it has been made possible for today's MLLMs to perceive information in texts, images, videos, etc., and solve complicated problems in physical environments \cite{achiam2023gpt,team2023gemini}. Along with the development of MLLMs, new benchmarks are continuously created to facilitate comprehensive and in-depth analysis of MLLMs \cite{yue2023mmmu,liu2023mmbench,fu2023mme,mvbench2023}.

However, it remains a great challenge to evaluate the MLLMs' long-video understanding (LVU) performances given the following limitations. Firstly, the majority of existing video understanding benchmarks are made up of short videos \cite{msrvtt2016,mvbench2023,seedbench2023,videobench2023,cvrr-2024}, whose lengths can be merely a few seconds. As a result, they are insufficient to reflect the MLLMs' long-video understanding capabilities. Secondly, there is a notable lack of diversity in both video genres and evaluation tasks. Existing benchmarks often concentrate on a single video type, such as egocentric videos \cite{ego4d2022, egoschema2023}, or focus on one specific task, like captioning \cite{msrvtt2016}. These limitations hinder comprehensive evaluation of LVU capabilities. Last but not least, many previous evaluation tasks are not properly designed for LVU, as they can be solved without using the complex information from long videos. For example, many questions are simply about one single frame in the long videos \cite{moviechat2023, zhang2024longva}. Besides, numerous others are about popular movies and celebrities \cite{llama-vid2023, videomme2024}, which can be answered directly by MLLMs based on the textual prompts.

Conceptually, MLLMs are expected to handle any type of long video and accomplish any related tasks. Therefore, the evaluation of LVU should emphasize two important properties: \textit{length} and \textit{diversity}. Furthermore, it is crucial that the evaluation tasks are specifically designed to leverage the complex information inherent in long videos, addressing the shortcomings of previous benchmarks. Based on these principles, we propose a novel benchmark called \textbf{MLVU} (\uline{M}ult-task \uline{L}ong \uline{V}ideo \uline{U}nderstanding Benchmark), which presents the following critical advantages.

\begin{table*}[t!]
\centering
\vspace{5pt}
\renewcommand{\arraystretch}{1.0} 
\begin{adjustbox}{max width=\textwidth}
\begin{tabular}{l>{\centering\arraybackslash}m{1.2cm}>{\centering\arraybackslash}m{1.2cm}>{\centering\arraybackslash}m{1.2cm}>{\centering\arraybackslash}m{1.2cm}>{\centering\arraybackslash}m{1.2cm}>{\centering\arraybackslash}m{1.5cm}>{\centering\arraybackslash}m{1.5cm}>{\centering\arraybackslash}m{1.5cm}>{\centering\arraybackslash}m{1.8cm}}
\toprule
\textbf{Benchmarks} & \textbf{\#Videos} & \textbf{\#QA Pairs} & \textbf{Len. (s)} & \textbf{Close-Ended} & \textbf{Open-Ended} & \textbf{Various Genres} & \textbf{Multi-Level} & \textbf{Multi-Dimension} & \textbf{Referring QA} \\
\midrule
NExT-QA~\cite{nextqa2021} & 1,000 & 8,564 & 39.5 & \checkmark & \checkmark & \checkmark & \crossmark & \crossmark & \crossmark \\
TVQA~\cite{tvqa2018} & 15,253 & 15,253 & 11.2 & \checkmark & \crossmark & \crossmark & \crossmark & \crossmark & \crossmark\\
MSRVTT-QA~\cite{msrvtt2016} & 2,900 & 72,821 & 15.2 & \checkmark & \crossmark & \crossmark & \crossmark & \crossmark  &\crossmark \\
MVBench~\cite{mvbench2023} & 3,641 & 4,000 & 16.0 & \checkmark & \crossmark & \checkmark & \crossmark & \crossmark & \crossmark \\
\midrule
Movie101~\cite{movie101-2023} & 101 & - & 6144 & \crossmark & \checkmark & \crossmark & \crossmark & \crossmark & \crossmark \\
EgoSchema~\cite{egoschema2023} & 5,063 & 5,063 & 180 & \checkmark & \crossmark & \crossmark & \crossmark & \crossmark & \crossmark \\
MovieChat-1K~\cite{moviechat2023} & 130 & 1,950 & 500 & \checkmark & \checkmark & \crossmark & \crossmark & \checkmark & \crossmark \\
Video-MME$^*$~\cite{videomme2024} & 900 & 2,700 & 1024 & \checkmark & \crossmark & \checkmark & \checkmark & \crossmark & \crossmark \\
LongVideoBench$^*$~\cite{wu2024longvideobench} & 3,763 & 6,678 & 473 & \checkmark & \crossmark & \checkmark & \checkmark & \crossmark & \checkmark \\
\midrule
\textbf{MLVU}  & 1,730 & 3,102 & 930 & \checkmark & \checkmark & \checkmark & \checkmark & \checkmark & \checkmark \\
\bottomrule
\end{tabular}
\end{adjustbox}
\vspace{-0.2cm}
\caption{Comparison of MLVU with existing benchmarks, including the number of videos (\textbf{\#Videos}), number of QA pairs (\textbf{\#QA pairs}), average video length (\textbf{Len.}), presence of \textbf{Close-Ended} tasks, presence of \textbf{Open-Ended} tasks, inclusion of various video genres (\textbf{Various Genres}), coverage of multiple duration levels (\textbf{Multi-Level}), inclusion of multiple dimensions of LVU tasks (\textbf{Multi-Dimension}), and questions involving local information with clear referring context rather than direct timestamps~\cite{moviechat2023} or well-known narrative elements~\cite{movienet2020,llama-vid2023} (\textbf{Referring QA}). The first block represents short video understanding benchmarks, and the second block represents long video understanding benchmarks. $^*$ denotes work concurrent with MLVU.} 
\vspace{-0.5cm}
\label{tab:comparison}
\end{table*}

\begin{itemize}
     \item \textbf{It makes a substantial extension for the video length.} MLVU is created based on long videos of diversified lengths, \textit{ranging from 3 minutes to 2 hours}. The average video length is about 15 minutes, which makes it much longer than most of the existing benchmarks. Additionally, each video is further segmented so that evaluation tasks can be created w.r.t. different video clips (e.g., summarization for the first 3 minutes, the first 6 minutes, and the entire duration of the video). Therefore, it is able to flexibly evaluate the MLLMs' performance across different video lengths. 

     \item \textbf{It encompasses a wide variety of video genres.} MLVU includes diverse real-world videos, such as movies, life records, and egocentric videos. Additionally, it features typical simulated videos like games and cartoons. This diversity allows for a comprehensive assessment of MLLMs' performance across various application scenarios.

     \item \textbf{It introduces diversified evaluation tasks tailored for LVU.} MLVU comprises 9 distinct tasks that collectively assess a wide range of MLLMs' LVU capabilities. On one hand, it includes both \textit{multiple-choice and open-ended generation} tasks, reflecting the models' performance in handling different task formats. On the other hand, some tasks are designed to leverage \textit{global information from entire videos}, while others require the use of \textit{specific local information from certain clips}. Moreover, all questions involving local information are annotated with unambiguous context, requiring MLLMs to accurately locate or infer the appropriate clips within long videos.
\end{itemize}

Table~\ref{tab:comparison} shows that MLVU provides a more comprehensive evaluation of LVU compared to existing and concurrent benchmarks. We extensively investigate 23 popular MLLMs with MLVU, which brings in several critical insights. Firstly, \textit{long-video understanding remains a technically challenging problem for the existing MLLMs}. While GPT-4o\footnote{https://openai.com/index/hello-gpt-4o/} achieves the leading performance in the experiment, it only attains an average score of 54.5\% in multi-choice tasks. All methods struggle with tasks requiring fine-grained information from entire videos, such as action counting, ordering, and summarization. 
Secondly, \textit{recent open-source long video MLLMs have made significant strides in LVU}~\cite{zhang2024longva,fei2024videoccam,shu2024videoxl}. These advancements have improved the models' capability to process extended visual sequences, thereby closing the gap with leading proprietary models in recent months.
Finally, \textit{the empirical results underscore influential factors in LVU}, such as the extension of context length, the improvement of image understanding ability, and the utilization of strong LLM-backbones. In addition to the benchmark's overall conclusion, individual tasks enable fine-grained analysis of MLLMs' performances in each specialized aspects. Therefore, we anticipate the benchmark to assist in improving MLLMs' long-video understanding capabilities by providing insights into their current strengths and weaknesses.

\vspace{-0.15cm}
\section{Related Work}
\vspace{-0.15cm}

\begin{figure*}[h!]
    \centering
    \vspace{-0.3cm}
    \includegraphics[width=1\textwidth]{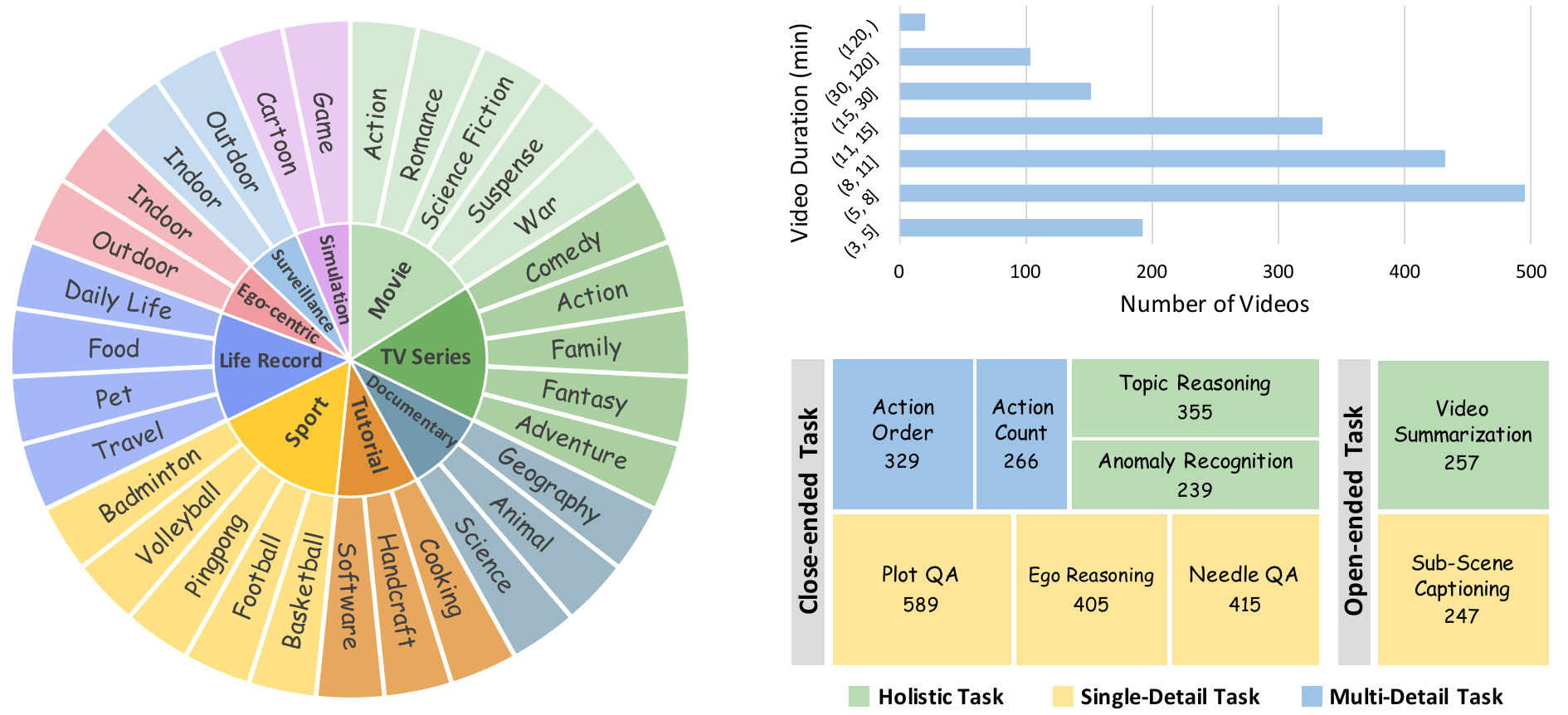}
    \vspace{-0.5cm}
    \caption{Statistical Overview of our MLVU benchmark. \textbf{Left:} Video genres included in MLVU; \textbf{Top Right:} Distribution of video duration; \textbf{Bottom Right:} Task types and their counts in MLVU.}    \label{fig:statistic}
    \vspace{-0.5cm}
\end{figure*}

\paragraph{Multimodal Large Language Models.} Multimodal large language models (MLLMs) have attracted significant interest from both academia and industry. Recent advancements in this field have been achieved by integrating LLM backbones with visual encoders and adapters, and fine-tuning the entire architecture through visual instruction tuning~\cite{llava2023, minigpt4-2023, internvl-1.5-2024}. Based on the same philosophy, MLLMs have been further developed for video processing using video instruction datasets and specialized video adapters~\cite{videollama, videochatgpt2023, videochat2023, mplug-owl-2023, videollava2023, mvbench2023}. However, most existing models are optimized for short videos, typically under one minute, due to the difficulty in establishing sufficient context for longer videos. 
To address this challenge, researchers have explored compact video representations or extended the context length of MLLMs. For instance, LLaMa-Vid~\cite{llama-vid2023} compresses each video frame into two tokens, enabling the model to handle videos several hours long. Methods like MovieChat~\cite{moviechat2023} and MA-LMM~\cite{malmm2024} introduce specialized memory components for recursive video processing. Furthermore, approaches such as LWM~\cite{liu2024world}, LongVA~\cite{zhang2024longva}, and Video-XL~\cite{shu2024videoxl} are designed to extend the context length of MLLMs, facilitating the processing of longer video inputs. Additionally, it is also explored to make selective usage of frames or clips from long videos based on retrievers or agents~\cite{R-VLM-2023, R2A2023, videoagent2024}. Despite these progresses, it remains an open problem for MLLMs to effectively handle long videos.

\vspace{-10pt}

\paragraph{Video Understanding Benchmarks.}
 
With the unprecedented interest in MLLMs, the creation of benchmarks for these models has become increasingly emphasized (as advanced by MMMU \cite{yue2023mmmu}, MME \cite{fu2023mme}, and many other pioneering works). In video understanding, the research community has made significant efforts as well, particularly for short videos. There are specialized benchmarks for temporal perception~\cite{activitynetqa2019, star2021}, action understanding~\cite{star2021, paxion2023}, video classification~\cite{kinetics-2017}, video reasoning~\cite{funqa2023, nextqa2021}, and video captioning~\cite{msrvtt2016, howto100m2019}. Recently, MVBench~\cite{mvbench2023} provides a comprehensive short-video benchmark to evaluate general capabilities via question-answering. 
For long video understanding, people seek to leverage long-form videos, like movies, to create benchmarks. For example, LLaMA-Vid~\cite{llama-vid2023} developed a movie question-answering dataset based on MovieNet~\cite{movienet2020}. Despite using long videos, many questions focus on well-known narrative elements, allowing them to be answered without analyzing the video's content. In contrast, MovieChat~\cite{moviechat2023} avoids specific character names or plot details in its questions. However, since each question provides a specific timestamp, the tasks can be reduced to short-video or image understanding problems. Beyond movies, there are task-specific benchmarks like EgoSchema~\cite{egoschema2023}, which presents video reasoning tasks using first-person footage from Ego4D~\cite{ego4d2022}. These specialized benchmarks, however, focus on a single aspect of MLLMs rather than offering a comprehensive analysis of long video understanding. Therefore, it is essential to develop a comprehensive benchmark with carefully designed tasks to effectively evaluate MLLMs' capabilities in understanding long videos. 

\section{MLVU: Multi-task Long Video Understanding Benchmark}
\label{sec:dataset}

In this section, we start with an overview of MLVU, which highlights its constitution and explains its values over the previous works. Then, we discuss how each evaluation task is constructed in MLVU. 

\subsection{Overview}
\label{sec:benchoverview}

MLVU is a multi-task benchmark consisting of 3,102 questions across 9 categories, specifically designed for long video understanding. It is divided into a dev set and a test set, containing 2,593 and 509 questions, respectively. The benchmark is distinguished by the following features.

\textbf{Diversified Video Categories.} MLVU offers a comprehensive collection of videos across various categories (Figure \ref{fig:statistic} Left). These include typical real-world videos such as movies, documentaries, TV series, egocentric videos, life records, sports, tutorials, and surveillance footage. Additionally, it features significant simulated videos from animated series and game videos.

\textbf{Substantial Extension of Video Length.} MLVU is made up of videos of diversified lengths, spanning from 3 min to more than 2 hours (Figure \ref{fig:statistic} Top Right). Besides, each video is further partitioned as incremental segments, e.g., the first 3 min, the first 6 min, and the entire video, where tasks are created for each individual segment. Thus, the MLLMs can be flexibly evaluated across different video lengths.

\textbf{Diversified Evaluation Tasks.} MLVU also provides a diverse array of evaluation tasks, which are closely aligned with the common visual capabilities of MLLMs, such as reasoning, captioning, recognition, perception, and summarization (Figure \ref{fig:statistic} Bottom Right). All the tasks are tailored for LVU. That is to say, the tasks need to be solved based on the in-depth understanding of video. Some of tasks are to examine whether the global information from the entire video can be effectively utilized (holistic LVU); while others focus on whether the MLLMs can make precise usage of proper local information within the long video (detail LVU). Additionally, both multi-choice and free-form generation tasks are included in MLVU, which help to examine MLLMs' capabilities in handling different task formats.

\subsection{Construction of MLVU}
\label{sec:MLVU-construction}

The evaluation tasks of MLVU can be categorized into three types: 1) \textit{holistic LVU}, which needs to be solved by making use of the global information from the entire video; 2) \textit{single-detail LVU}, which relies on leveraging one critical plot within the long video; and 3) \textit{multi-detail LVU}, which necessitates the joint utilization of multiple plots within the long video. The construction process of MLVU is discussed w.r.t the above three categories. To facilitate the discussion, we define \textit{ULVC} (Universal Long Video Collection) as the universal collection of long videos from various sources (more details about ULVC are presented in \Cref{appendix:ulvc}).

\subsubsection{Holistic LVU}
\begin{figure*}[thp]
    \centering
    \includegraphics[width=0.86\textwidth]{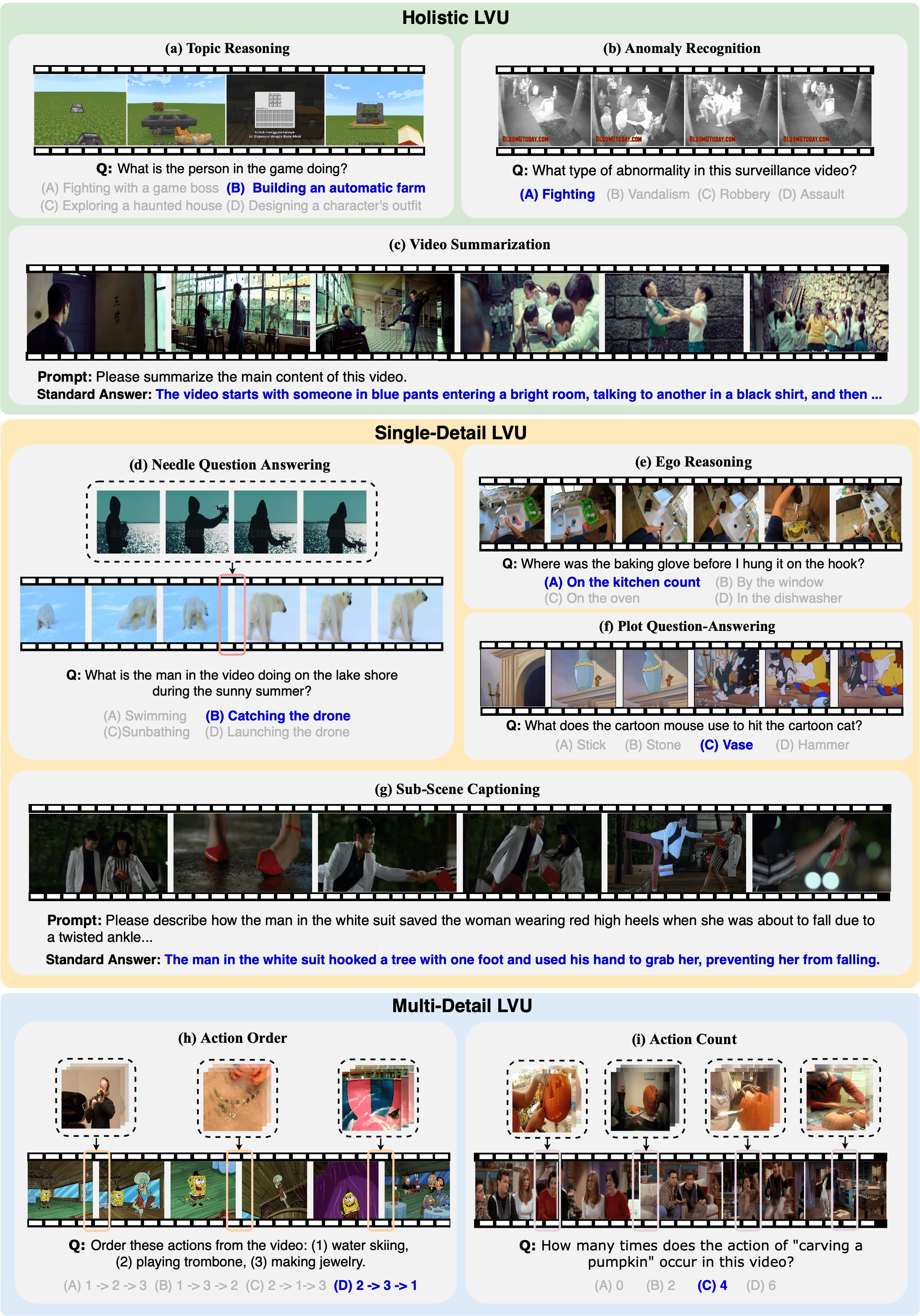}
    \caption{Examples of MLVU. There are nine tasks designed to evaluate the \textit{holistic}, \textit{single-detail}, and \textit{multi-detail LVU} capabilities of MLLMs. The MLLMs are asked to solve the problem (with the ground-truth answers marked in blue) based on the long video input and textual prompt. For multiple-choice questions, we set 4 candidates in the dev set and 6 candidates in the test set.} 
    \label{fig:pipeline}
\end{figure*}

\textbf{Topic Reasoning (TR).} The topic reasoning task requires MLLMs to respond to questions about the principal subject of a long video, as shown with Figure \ref{fig:pipeline} (a). This includes elements such as the video's genre, pivotal events, or primary settings. 
All questions and answers undergo manual annotation\footnote{Detailed information and annotation guidelines for annotators are presented in \Cref{appendix:anno-detail}.}, resulting in a total of 355 questions. TR tasks are formatted as multiple-choice questions, with the model's performance assessed based on accuracy.

\noindent\textbf{Anomaly Recognition (AR).} The anomaly recognition task involves identifying the anomalous behavior within a surveillance footage (Figure \ref{fig:pipeline} b). We leverage the surveillance video clips from UCF Crime dataset \cite{ucfcrime-2018} for this task. The selected video clips are longer than three minutes. We create 239 questions based on the original annotations provided by the dataset. The AR task is also conducted in the multiple-choice format, whose performance is measured by accuracy.

\noindent\textbf{Video Summarization (VS).} This task requires MLLMs to summarize the key events in a long video (Figure \ref{fig:pipeline} c). We select the narrative-rich videos from ULVC for this task, including movies, TV series, documentaries, life records, and animated series. There are 257 selected videos in total, whose summaries are manually annotated. During evaluation, the MLLMs are prompted with "Please summarize the main content of this video". We employ GPT-4 to assess the generated summaries by comparing with the annotation results. Details about annotation and evaluation are presented in Appendix \ref{sub-appendix:anno-detail-VS} and \ref{sub-appendix:evaluation_metric}.

\subsubsection{Single-Detail LVU}

\textbf{Needle Question-Answering (NQA).} Needle-In-the-Haystack-Search (NIHS) is a popular evaluation task for long-context LLM \cite{liu2024lost}. Taking the inspiration from NIHS, we create Needle Question-Answering (NQA), shown as Figure \ref{fig:pipeline} (d). In this task, the MLLM is required to answer a question related to a specific segment (referred as \textit{needle}) within a long video (referred as \textit{background video}). The needles are short video clips sampled from WebVid~\cite{webvid-2021} and Clevrer~\cite{clevrer-2019}, while the background videos are sampled from our ULVC. The needle is randomly inserted into the background video, where a question-answer pair is annotated. By incorporating necessary details, the question can always correspond to the needle without ambiguity. During evaluation, the MLLM needs to infer the location of the needle based on the details provided in the question, and solve the problem on top of the needle's information. The NQA task is structured as multiple-choice, whose performance is measured by accuracy.

\noindent\textbf{Ego Reasoning (ER)}. Ego-centric videos capture a series of consecutive actions from a first-person perspective. The MLLM needs to reason for a question about a specific behavior in the video, e.g., predicting for the event which is correlated or satisfies a certain causal relationship with the behavior (\Cref{fig:pipeline} e). Both videos and QA annotations are collected from the NLQ task of Ego4D \cite{ego4d2022}. The ER task is structured as multiple-choice, with a total of 405 questions created for this task.

\noindent\textbf{Plot Question-Answering (PQA)}. In this task, the MLLM needs to reason for questions about a plot in a narrative video, shown as \Cref{fig:pipeline} (f). The video is sampled from the movies, TV series, and animated series in our ULVC. There are 589  question-answer pairs created by manual annotation. During annotation, the human annotators are asked to only provide necessary details about the plot but not to suggest any objective hints, e.g., the two characters in the example video are referred as cat and mouse, rather than Tom and Jerry. Therefore, it can prevent the question from being short-cut by the MLLM's common-sense knowledge (more details about PQA can be found in the Appendix~\ref{sub-appendix:anno-detail-PQA}).

\noindent\textbf{Sub-Scene Captioning (SSC).} In this task, the MLLM needs to generate the caption for a sub-scene in a long video. The long videos in SSC are sampled from the Movie101 dataset \cite{movie101-2023}, while the questions and answers are manually annotated. During annotation, the human annotator is asked to provide a detailed description for the sub-scene as the ground-truth answer. Besides, they need to offer necessary clues in their questions such that the referred sub-scenes can be identified without ambiguity. During evaluation, we employ GPT-4~\cite{achiam2023gpt} to measure the quality of caption in comparison with the ground-truth. Details about annotation and evaluation are presented in Appendix~\ref{sub-appendix:anno-detail-SSC} and \ref{sub-appendix:evaluation_metric}.

\subsubsection{Multi-Detail LVU}

\textbf{Action Order (AO).} In this task, the MLLM needs to predict the right order for a sequence of actions (\Cref{fig:pipeline} h). The actions are presented by short video clips, called \textit{probes}. The probes are formulated in two different ways. One is made up of clips from the Kinetics dataset \cite{kinetics-2017}, where each clip represents a distinct action. The other one is from the consecutive clips of an action in the ActivityNet-Caption dataset~\cite{activitynet-caption2017}. The probes are inserted into a long \textit{background} video, which is sampled from ULVC. There are 329 AO questions in total. The task is structured as a multiple-choice prblem, where the right order is selected from the misleading options provided by the annotator.

\noindent\textbf{Action Count (AC).} This task requires the MLLM to count the occurrences of an action within a long video (\Cref{fig:pipeline} i). Each action corresponds to multiple short \textit{probe} clips sampled from the Kinetics dataset \cite{kinetics-2017}. The probes of an action are inserted into a long \textit{background} video sampled from ULVC. We also perform manual examination to ensure that the inserted action does not exist in the original background video. A total of 266 evaluation instances have been created. The AC task is structured as a multiple-choice problem, with performance measured by accuracy.

\renewcommand{\arraystretch}{1.2}
\begin{table*}[t]
\small
\resizebox{\linewidth}{!}{
\begin{tabular}{lccccccccccccc}
\toprule
\specialrule{0em}{0.3pt}{0.3pt}
\multirow{2}{*}{\textbf{Methods}} & \multirow{2}{*}{\textbf{Date}} & \multirow{2}{*}{\textbf{Input}}  & \multicolumn{3}{c}{\textbf{Holistic}} & \multicolumn{4}{c}{\textbf{Single Detail}}    & \multicolumn{2}{c}{\textbf{Multi Detail}} & \multirow{2}{*}{\textbf{M-Avg}} & \multirow{2}{*}{\textbf{G-Avg}} \\ 
\specialrule{0em}{0.3pt}{0.3pt}
\cmidrule(r){4-6} \cmidrule(r){7-10} \cmidrule(r){11-12}
\specialrule{0em}{0.1pt}{0.1pt}
&~&~& TR   &AR & VS$^{*}$  & NQA &ER &PQA
& SSC$^{*}$   & AO &AC \\     
\specialrule{0em}{0.3pt}{0.3pt}
\hline
\rowcolor[HTML]{eff0f1}Full mark &-- &-- &100 &100  &10  &100  &100  &100  &10  &100  &100  &100  &10    \\
\rowcolor[HTML]{eff0f1}Random &-- &-- &16.7 &16.7  &--  &16.7  &16.7  &16.7  &--  &16.7  &16.7  &16.7  &--    \\

\hline
 \rowcolor[HTML]{E3F8F8}\multicolumn{14}{l}{\gray{\textit{\textbf{Image MLLMs}}}}\\
 \rowcolor[HTML]{E3F8F8}  Otter-I~\cite{otter2023} &2023-05 &16 frm  & 17.6 & 17.9 &2.03   & 16.7  & 17.0  & 18.0 &3.90 & 15.7 & 16.7 & 17.1 & 2.97 \\
          \rowcolor[HTML]{E3F8F8}LLaVA-1.6~\cite{llava2023} &2024-01 &16 frm  & 63.7 & 17.9  & 2.00  & 13.3  & 26.4  & 30.0 &4.20   & 21.4 & 16.7 & 27.1 & 3.10 \\
         \rowcolor[HTML]{E3F8F8} InternVL-2~\cite{internvl-1.5-2024} &2024-07 &16 frm & 85.7 & 51.3 & 2.55  & 48.3  & 47.2  & 52.0  & 5.25  & 32.9 & 15.0 & 47.5 & 3.90   \\
         \rowcolor[HTML]{E3F8F8} Claude-3-Opus$^\dag$~\cite{Claude3} &2024-03 &16 frm  & 53.8 & 30.8 & 2.83  & 14.0  & 17.0  & 20.0  & 3.67  & 10.0 & 6.7  & 21.8 & 3.25  \\
        \rowcolor[HTML]{E3F8F8} Qwen-VL-Max$^\dag$~\cite{qwenvl-2023} &2024-01 &16 frm  & 75.8 & 53.8 & 3.00  & 15.0  & 26.4  & 4.84  & 20.0  & 20.7 & 11.7 & 32.2 & 3.92  \\
        \hline
         \rowcolor[HTML]{FFF5F5}\multicolumn{14}{l}{\gray{\textit{\textbf{Short Video MLLMs}}}} \\
        \rowcolor[HTML]{FFF5F5}Otter-V~\cite{otter2023} &2023-05  &16 frm  & 16.5 & 12.8 & 2.18  & 16.7  & 22.6  & 22.0  & 4.20   & 12.9 & 13.3 & 16.7 & 3.19  \\
        \rowcolor[HTML]{FFF5F5}mPLUG-Owl-V~\cite{mplug-owl-2023} &2023-04  &16 frm  & 25.3 & 15.4 & 2.20  & 6.7   & 13.2  & 22.0  & 5.01  & 14.3 & 20.0 & 16.7 & 3.61 \\
         \rowcolor[HTML]{FFF5F5}VideoChat~\cite{videochat2023} &2023-05 &16 frm   & 26.4 & 12.8 & 2.15  & 18.3  & 17.0  & 22.0  & 4.90  & 15.7 & 11.7 & 17.7 & 3.53  \\
        \rowcolor[HTML]{FFF5F5}Video-LLaMA-2~\cite{videollama} &2024-08  &16 frm  & 52.7 & 12.8 & 2.23  & 13.3  & 17.0  & 12.0  & 4.87  & 15.7 & 8.3  & 18.8 & 3.55  \\
         \rowcolor[HTML]{FFF5F5}VideoChat2-HD~\cite{mvbench2023} &2024-06 &16 frm  & 74.7 & 43.6 & 2.83  & 35.0  & 34.0  & 30.0  & 5.14  & 21.4 & 23.3 & 37.4 & 3.99  \\
          \rowcolor[HTML]{FFF5F5}Video-LLaVA~\cite{videollava2023}  &2023-11 &8 frm  & 70.3 & 38.5 & 20.9  & 2.30  & 26.4  & 26.0  & 5.06  & 20.0 & 21.7 & 29.3 & 3.68  \\
            \rowcolor[HTML]{FFF5F5}ShareGPT4Video~\cite{chen2024sharegpt4video} &2024-05 &16 frm  & 73.6 & 25.6 & 2.53  & 31.7  & 45.3  & 38.0  & 4.72  & 17.1 & 8.3  & 34.2 & 3.63  \\
           \rowcolor[HTML]{FFF5F5}VideoLLaMA2~\cite{cheng2024videollama2} &2024-06 &16 frm   & 80.2 & 53.8 & 2.80  & 36.7  & 54.7  & 54.0  & 5.09  & 42.9 & 16.7 & 48.4 & 3.95  \\
        \hline
        \rowcolor[HTML]{F1F6EC}\multicolumn{14}{l}{\gray{\textit{\textbf{Long Video MLLMs}}}}\\
         \rowcolor[HTML]{F1F6EC}MovieChat~\cite{moviechat2023} &2023-07 &2048 frm   & 18.7 & 10.3 & 2.30  & 23.3  & 15.1  & 16.0  & 3.24  & 17.1 & 15.0 & 16.5 & 2.77    \\
         \rowcolor[HTML]{F1F6EC}Movie-LLM~\cite{moviellm2024} &2024-03 &1 fps  & 27.5 & 25.6 & 2.10  & 10.0  & 11.3  & 16.0  & 4.93  & 20.0 & 21.7 & 18.9 & 3.52   \\
        \rowcolor[HTML]{F1F6EC}LLaMA-VID~\cite{llama-vid2023} &2023-11 &1 fps  & 20.9 & 23.1 & 2.70  & 21.7  & 11.3  & 16.0  & 4.15  & 18.6 & 15.0 & 18.1 & 3.43  \\
        \rowcolor[HTML]{F1F6EC}MA-LMM~\cite{malmm2024} &2024-04 &1000 frm & 44.0 & 23.1 & 3.04  & 13.3  & 30.2  & 14.0  & 4.61  & 18.6 & 13.3 & 22.4 & 3.83   \\
          \rowcolor[HTML]{F1F6EC}
        MiniGPT4-Video~\cite{minigpt4video-2024} &2024-04 &90 frm  & 64.9 & 46.2 & 2.50  & 20.0  & 30.2  & 30.0  & 4.27  & 15.7 & 15.0 & 31.7 & 3.39   \\
          \rowcolor[HTML]{F1F6EC}
        LongVA~\cite{zhang2024longva} &2024-06 &256 frm  & 81.3 & 41.0 & 2.90  & 46.7  & 39.6  & 46.0  & 4.92  & 17.1 & 23.3 & 42.1 & 3.91   \\
          \rowcolor[HTML]{F1F6EC}
        Video-CCAM~\cite{fei2024videoccam} &2024-08 &96 frm  & 79.1 & 38.5 & 2.65  & 45.0  & 52.8  & 56.0  & 4.49  & 24.3 & 26.7 & 46.1& 3.57   \\
          \rowcolor[HTML]{F1F6EC}
        Video-XL~\cite{shu2024videoxl} &2024-09 &256 frm  & 78.0 & 28.2 & 3.40  & 50.0  & 41.5  & 46.0  & 5.02  & 48.6 & 31.7 & 46.3 & 4.21   \\
          \rowcolor[HTML]{F1F6EC}
        LLaVA-Onevision~\cite{li2024llavaonevision} &2024-08 &32 frm  & 83.5 & 56.4 & 3.75  & 46.7  & 58.4  & 58.0  & 5.09  & 35.7 & 23.3 & 51.7 & 4.42   \\
        \rowcolor[HTML]{F1F6EC}GPT-4o$^\dag$~\cite{gpt4o}  &2024-05 &0.5 fps  & 83.7 & 68.8 & 4.94  & 42.9  & 47.8  & 57.1  & 6.80  & 46.2 & 35.0 & 54.5 & 5.87  \\
\bottomrule
 \end{tabular}
}
\vspace{-0.2cm}
\caption{The overall performances on MLVU test set, including the holistic LVU tasks, the single-detail LVU tasks, and multi-detail LVU tasks. Date: the release date of the MLLM. M-Avg: the average performance of multiple-choice tasks; G-Avg: the average performance of generation tasks (marked by $*$). Two input strategies are used by the MLLMs in evaluation: Uniform Sampling (\textbf{N frm}), which evenly samples N frames from the video; Frame Rate Sampling (\textbf{N fps}), which samples N frames per second. $\dag$ denotes proprietary models.} 
\vspace{-0.5cm}
\label{tab:overall:test}
\end{table*}

\section{Experiments and Analysis}
\label{sec:experiments}

\subsection{Settings}

We conduct a comprehensive investigation of 23 MLLMs using our MLVU benchmark, encompassing both open-source and proprietary models. The experimental MLLMs are divided into three categories: \textbf{\textit{1) Image MLLMs}}, primarily fine-tuned using image-related instructions; \textbf{\textit{2) Short Video MLLMs}}, fine-tuned with short-video related instructions; and \textbf{\textit{3) Long Video MLLMs}}, optimized for long-video understanding capability. For Image MLLMs, we leverage their multi-image inference capabilities to process segmented frames from original videos. For Video MLLMs, we employ either a uniform sampling strategy or a frame rate sampling strategy for video processing. All models are evaluated based on their official implementations or available APIs, with evaluations conducted in a zero-shot manner. More details about the evaluation are provided in Appendix \ref{appendix:evaluation}.

\subsection{Main Results}

The overall evaluation results for all investigated MLLMs in the MLVU test set are shown in \Cref{tab:overall:test} (with dev set results in Appendix \ref{appendix:dev-result}). Individual performances are reported for each task, while average performances are provided for multiple-choice (M-Avg) and generation tasks (G-Avg). From the results, we derive three primary conclusions:

\textbf{1) The proprietary model GPT-4o~\cite{gpt4o} achieves optimal performance in our benchmark.} It leads in multiple-choice tasks with an M-Avg of 54.5\%(within 0-100\%) and excels in generation tasks with a G-Avg of 5.87 (within 0.0-10.0), outperforming all other methods.

\textbf{{2) Recent advances in LVU have achieved significant progress, and the gap between open-source long video MLLMs and GPT-4o on close-ended tasks is narrowing.}} Before June 2024, the best open-source long video MLLMs, MiniGPT4-Video~\cite{minigpt4video-2024}, lagged significantly behind GPT-4o. However, recent models~\cite{zhang2024longva, fei2024videoccam, shu2024videoxl, li2024llavaonevision} have made substantial progress. For instance, LLaVA-Onevision trails GPT-4o by only 2.8\% in M-Avg. These models have improved their ability to handle long visual sequences, achieving significant advancements in single-detail (e.g., NQA) and multi-detail (e.g., AC) tasks compared to previous open-source models.

\textbf{{3) Existing methods still struggle to handle most tasks in our benchmark.}} For instance, GPT-4o only achieves 42.9\% in the needle question-answering (NQA) task. In contrast, analogous tasks in the text domain, such as NIHS (Needle-In-the-Haystack-Search) and Passkey Retrieval, are effectively handled by many existing long LLMs \cite{fu2024data,inftybench-2024}. Additionally, GPT-4o shows even less reliability in tasks like ego-reasoning (ER), action ordering (AO), and action count (AC), with most baseline methods performing even worse. 
These observations indicate that long-video understanding remains a significant challenge for today's MLLMs.

In addition to the primary conclusions from the overall performances, we can also make the following interesting observations about the individual tasks.

\textbf{{4) The close-ended holistic tasks present much higher differentiation than other tasks.}} These tasks, i.e., topic reasoning (TR) and anomaly recognition (AR), show significant variance in performance across different models. Proprietary MLLMs, like GPT-4o, and superior open-source models, such as InternVL-2~\cite{internvl-1.5-2024}, VideoLLaMA2~\cite{cheng2024videollama2}, and LLaVA-OneVision~\cite{li2024llavaonevision}, can accurately solve these problems. Meanwhile, many other popular MLLMs still fail to generate meaningful performances. Since these tasks only require an overall understanding of long videos, they can serve as a preliminary indicator of MLLMs' long video understanding (LVU) ability.

\textbf{{5) It's challenging to deal with tasks that require nuanced understanding of multiple details.}} Although several MLLMs can handle single-detail LVU tasks to some extent, their performances suffer from catastrophic degradation when addressing multi-detail LVU tasks. Most methods, except for GPT-4o and Video-XL~\cite{shu2024videoxl}, fail entirely in action order (AO) and action count (AC) tasks. Additionally, most approaches struggle with summarization tasks, which require recalling multiple nuanced details from long videos.

As a brief conclusion, although today's MLLMs can deal with some preliminary LVU tasks, it remains a tough challenge to achieve an in-depth understanding of nuanced information within long videos.

\subsection{Further Analysis}
\label{sec-exp-detail-analysis}

\renewcommand{\arraystretch}{1.2}

\begin{table*}[tp]
    \centering
    \setlength\tabcolsep{2.5pt}
    \resizebox{0.85\linewidth}{!}{
        \begin{tabular}{l c l||l c l||l c l}
        \Xhline{1.0pt}
        \multicolumn{3}{c||}{\textbf{Impact of Context Length}} & \multicolumn{3}{c||}{\textbf{Impact of IU}} &   \multicolumn{3}{c}{\textbf{Impact of LLM}} \\
        \Xhline{0.8pt}
        Model & Context Len. &M-Avg & Model & MMMU (Val) &M-Avg & Model & LLM & M-Avg\\
        \Xhline{0.8pt}
        \multirow{2}{*}{MGV} & 16 & 24.2 & Otter-I & 32.2 &17.1 & \multirow{2}{*}{VLM2} &Vicuna-7B &13.3 \\
        ~ & 90 & 31.7\darkGreen{$\uparrow$7.5} & LLaVA-1.6 & 35.8 &27.1\darkGreen{$\uparrow$10.0}  &~ & Vicuna-13B & 18.8\darkGreen{$\uparrow$5.5}    \\
        \hline
        \multirow{2}{*}{GPT-4o} & 16 & 45.8 & GPT-4V& 58.1& 43.3 & \multirow{2}{*}{MGV} & LLaMA-7B & 20.6 \\
        ~ & 256 & 54.5\darkGreen{$\uparrow$8.7}  & GPT-4o & 63.8 &45.8\darkGreen{$\uparrow$2.5}  & ~ & Mistral-7B &31.7 \darkGreen{$\uparrow$11.1}  \\
        \Xhline{1.0pt}
        \end{tabular}
    }
    \vspace{-0.25cm}
    \caption{Detailed discussions about the impact from context length, image understanding (IU) ability, and LLM Backbone. For the IU impact experiment, we used 16-frame uniform sampling for both GPT-4V and GPT-4o. MGV: MiniGPT4-Video, VLM2: Video-LLaMA-2.} 
    \label{tab:ablations}
    \vspace{-0.5cm}
\end{table*}

\textbf{{6) Longer videos are more challenging for MLLMs. }}We evaluate MLLMs' performances across various video lengths. For this purpose, we introduce a derivative dataset alongside MLVU, called \textit{MLVU Time-ladder}. In this dataset, the same kinds of evaluation tasks are created for videos of variant lengths, including 180s, 360s, and 600s (more details presented in \Cref{appendix:Time-Ladder}). As shown in Figure \ref{fig:length}, the performances of all models tend to decline as the video length grows, which indicates that the existing MLLMs' LVU abilities are severely constrained by the video length. Moreover, the short video model Video-LLaMA-2~\cite{videollama} maintains a certain level of LVU ability at 3 minutes, but its performance approaches random results at 10 minutes.

\begin{figure}[t]
    \centering
    \includegraphics[width=0.45\textwidth]{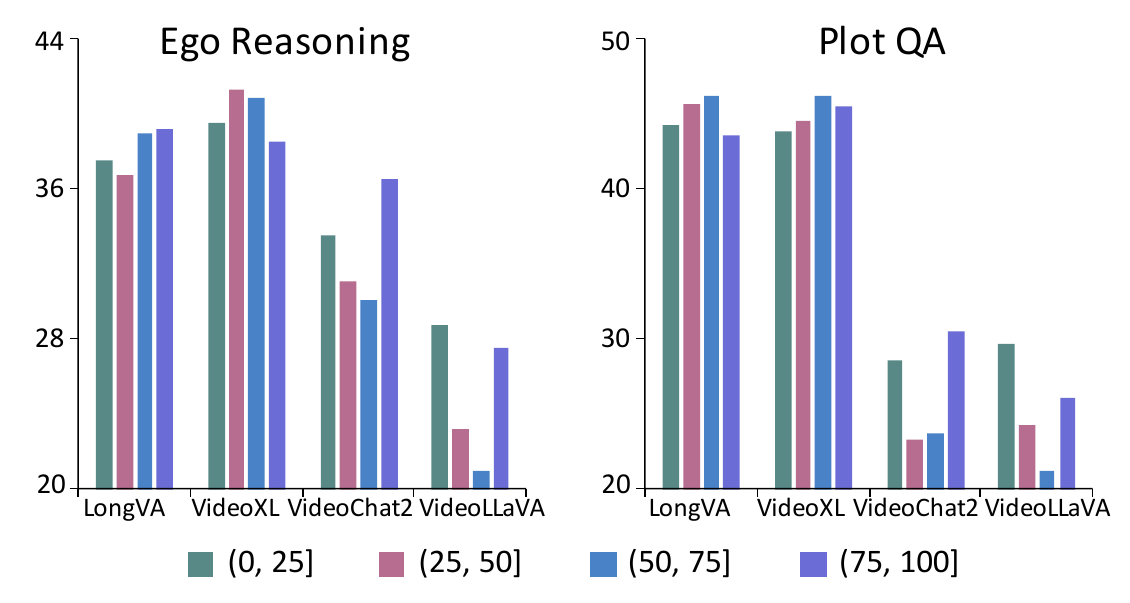}
    \vspace{-0.3cm}
    \caption{Model performance across different referring clip positions, spanning from the beginning to the end of the entire video.} 
    \vspace{-0.6cm}
    \label{fig:needle_position}
\end{figure}

\textbf{7) The performance of recent advanced long video MLLMs remains robust regardless of the position of the referring clip within the long video.} In single-detail tasks, the referring clip denotes the specific segment of the long video that is referenced or inferred to answer a question. As shown in Figure \ref{fig:needle_position}, we categorize clip positions into four intervals and assess model performance on two single-detail tasks: ego reasoning (ER) and plot question-answering (PQA). Recent long video MLLMs, such as LongVA~\cite{zhang2024longva} and Video-XL~\cite{shu2024videoxl}, maintain consistent performance regardless of the referring clip's position within the video. Conversely, short video MLLMs are more sensitive to clip location. This indicates that recent advancements in long video MLLMs enhance both reliable clue retrieval and effective reasoning from extended visual sequences.

\begin{figure}[t]
    \centering
    \includegraphics[width=0.45\textwidth]{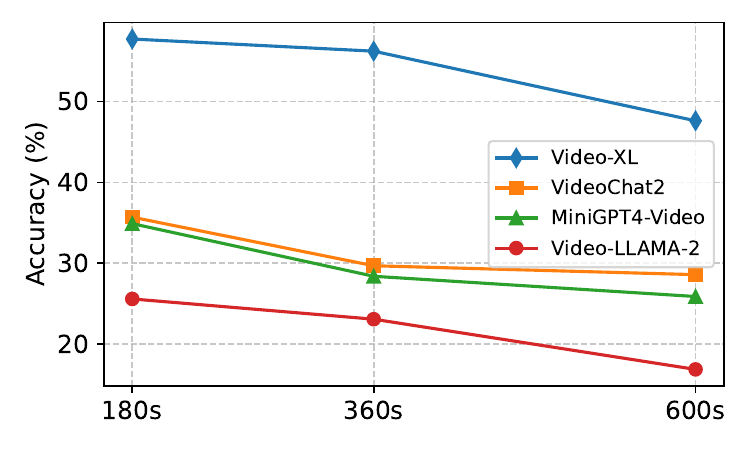}
    \vspace{-0.2cm}
    \caption{Experimental performance on varying video lengths. The evaluated metric is the average accuracy across five multiple-choice tasks involving local information: NQA, ER, PQA, AC, and AO.}  
    \label{fig:length}
    \vspace{-0.3cm}
\end{figure}

\begin{figure}[t]
    \centering
    \includegraphics[width=0.45\textwidth]{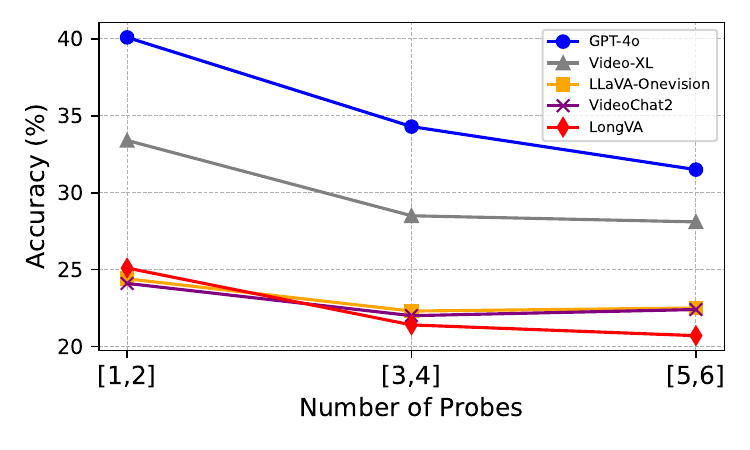}
    \vspace{-0.2cm}
    \caption{Model performance on the action count (AC) task in relation to the number of probes.}  
    \vspace{-0.5cm}
    \label{fig:multi_detail_num}
\end{figure}

\textbf{8) The challenge of multi-detail tasks increases with the number of details.} We analyzed model performance on the action count (AC) task by grouping questions based on the number of probes (which correspond to details) and evaluating the average performance within these groups. As shown in Figure~\ref{fig:multi_detail_num}, performance significantly declines across all models as the number of probes increases. This indicates that current MLLMs face substantial difficulties comprehending and processing multiple details simultaneously, highlighting a critical area for future improvement in long video understanding capabilities.

\textbf{{9) Context Length, Image-Understanding ability, and the choice of LLM Backbones are key factors in LVU performance.}} As shown in Table \ref{tab:ablations}, we conducted ablation experiments on several factors affecting MLLMs, using M-Avg as the evaluation metric. First, we examined the models' handling of different context lengths. Specifically, we increased MiniGPT4-Video's input from 16 to 90 frames and GPT-4o's input from 16 to 256 frames (as shown on the left side of Table \ref{tab:ablations}). Both models showed consistent performance improvements with longer input lengths. To assess the impact of image understanding (IU) capabilities, we referred to the results from MMMU \cite{yue2023mmmu} (presented in the middle of Table \ref{tab:ablations}). It is evident that MLLMs' LVU performance generally aligns with their IU performance in MMMU. Finally, we compared MLLMs using different backbones (depicted on the right side of Table \ref{tab:ablations}). The findings indicate that LVU performance improves with larger (Vicuna-13B vs. Vicuna-7B) and more advanced backbones (Mistral-7B vs. Llama-2-7B). These observations indicate that LVU is the result of multiple complex factors, with the ability to perceive longer videos and effectively utilize the perceived information being crucial for the improvement of LVU. 

\vspace{-0.2cm}
\section{Conclusion}
\vspace{-0.2cm}
\label{conclusion}
This paper presents MLVU, a novel benchmark for the assessment of long video understanding. With several critical innovations: the substantial extension of video lengths, the inclusion of various video genres, and the development of diversified LVU-oriented evaluation tasks, the new benchmark is able provide a comprehensive and in-depth analysis for MLLMs' long-video understanding performance. The empirical study on MLVU reveals LVU remains a technically challenging problem for today's state-of-the-art MLLMs. Future advancements may call for the joint optimization of complex factors, such as context length, image understanding ability, and even LLM backbones. We anticipate this benchmark will facilitate future research in long-video understanding of MLLMs. 

{
    \small
    \bibliographystyle{ieeenat_fullname}
    \bibliography{main}
}

\clearpage
\setcounter{page}{1}
\maketitlesupplementary

\appendix 
\setcounter{figure}{0}
\setcounter{table}{0}

\section{Overview of Appendix}
\begin{itemize}
    \item \ref{appendix:dev-result}: \textbf{Evaluation Results on MLVU Dev Set}
    \item \ref{appendix:ulvc}: \textbf{Collecting Details of our Universal Long Video Collection (ULVC)}
    \item  \ref{appendix:Time-Ladder}: \textbf{Details of the MLVU Time-Ladder}
    \item \ref{appendix:detail-of-dev-test} \textbf{Detailed Division of Dev and Test Sets in MLVU}
    \item \ref{appendix:anno-detail}: \textbf{Annotation Details of MLVU}
    \item \ref{appendix:evaluation}: \textbf{Details of Baselines and the Evaluation Process}.
    \item \ref{appendix:rag}: \textbf{Explorations of Video Retrieval Augmented Generation}
    \item \ref{appendix:more-mlvu-example}: \textbf{More Visualized Examples of MLVU}
\end{itemize}

\section{Evaluation Results on MLVU Dev Set}
\label{appendix:dev-result}
The evaluation results of the baselines on the MLVU dev set are detailed in Table~\ref{tab:overall}. Notably, the multiple-choice questions in the MLVU dev set present four options, whereas the MLVU test set offers six, making the latter more challenging and discriminative.

\section{Collecting Details of our Universal Long Video Collection (ULVC)}
\label{appendix:ulvc}
In the initial stage of our Multi-task Long Video Understanding (MLVU) benchmark creation, we first collected long-form videos from a variety of sources to form our Universal Long Video Collection (ULVC). The entirety of the long videos incorporated into our MLVU benchmark were selected, edited, or synthesized from ULVC. 

Specifically, our ULVC includes a diverse set of 986 long videos. This collection features 168 movies from the Movie101~\cite{movie101-2023} and MovieChat~\cite{moviechat2023} datasets, along with 60 documentaries from MovieChat~\cite{moviechat2023}. It also contains 65 game videos from MineDojo~\cite{minedojo2022}, 239 surveillance videos from UCF-Crime~\cite{ucfcrime-2018}, and 100 ego-centric videos from Ego4D~\cite{ego4d2022}. Additionally, we independently collected 72 cartoons, 92 TV series, 60 tutorial videos, 60 sports videos, and 70 life records.

It's important to clarify that the quantity of videos in the ULVC does not directly correspond to the number of videos and questions in our MLVU benchmark, which are 1,730 and 3,102 respectively. For example, a two-hour movie from the ULVC might be utilized in its entirety for the Sub-Scene Captioning task, or it could be segmented into several approximately 10-minute clips for the Video Summarization task, or even used as a background video for synthetic video generation. Moreover, a single video could be annotated with multiple questions simultaneously.

\section{Details of the MLVU Time-Ladder}
\label{appendix:Time-Ladder}
As discussed in Section \ref{sec:benchoverview}, most tasks in our MLVU are subject to segment-level annotation. This approach provides us with the flexibility to adjust the length of the video without requiring additional human annotators. Building on this strategy, as mentioned in Section \ref{sec-exp-detail-analysis}, we have generated a derivative dataset, \textit{MLVU Time-Ladder}, which includes videos of varying durations - specifically 3, 6, and 10 minutes. This dataset allows us to investigate how video duration impacts LVU task difficulty.

Specifically, during the annotation process of the VS task, we guided annotators to delineate the summarization in accordance with the initial 3 and 6-minute segments. For the PQA and SSC tasks, we requested annotators to identify the segments within the extended video where the pertinent answers are located. In the case of the ego reasoning task, the Ego4D dataset~\cite{ego4d2022} already comprises the intervals where the answers reside. Lastly, for the synthetic tasks of NQA, AO, and AC, we possess the capability to directly generate the necessary video lengths.

\section{Detailed Division of Dev and Test Sets in MLVU}
\label{appendix:detail-of-dev-test}
Our MLVU comprises a total of 3,102 questions, divided into a dev set with 2,593 questions and a test set with 509 questions. We present the detailed distribution of questions for each task in Table~\ref{tab:division}.

\begin{table}[ht]
    \centering
    \begin{tabular}{lccc}
        \toprule
        Task & Dev & Test & Total \\
        \midrule
        Topic Reasoning     & 264    &  91 & 355 \\
        Anomaly Recognition & 200    &  39 & 239 \\
        Video Summarization & 217    &  40  & 257 \\
        \midrule
        Needle QA           & 355    &  60   & 415 \\
        Ego Reasoning       & 352    &  53  & 405 \\
        Plot QA             & 539    &  50  & 589 \\
        Sub-Scene Captioning & 201    & 46  & 247 \\
        \midrule
        Action Order         & 259    & 70  & 329 \\
        Action Count         & 206    & 60  & 266 \\
        \bottomrule
    \end{tabular}
    \caption{Detailed Distribution of Questions in the MLVU Dataset Across Dev and Test Sets for Each Task.}
    \label{tab:division}
\end{table}

\renewcommand{\arraystretch}{1.2}
\begin{table*}[t]
\small
\resizebox{\linewidth}{!}{
\begin{tabular}{lccccccccccccc}
\toprule
\specialrule{0em}{0.3pt}{0.3pt}
\multirow{2}{*}{\textbf{Methods}} & \multirow{2}{*}{\textbf{Date}} & \multirow{2}{*}{\textbf{Input}}  & \multicolumn{3}{c}{\textbf{Holistic}} & \multicolumn{4}{c}{\textbf{Single Detail}}    & \multicolumn{2}{c}{\textbf{Multi Detail}} & \multirow{2}{*}{\textbf{M-Avg}} & \multirow{2}{*}{\textbf{G-Avg}} \\ 
\specialrule{0em}{0.3pt}{0.3pt}
\cmidrule(r){4-6} \cmidrule(r){7-10} \cmidrule(r){11-12}
\specialrule{0em}{0.1pt}{0.1pt}
&~&~& TR   &AR & VS$^{*}$  & NQA &ER &PQA
& SSC$^{*}$   & AO &AC \\     
\specialrule{0em}{0.3pt}{0.3pt}
\hline
\rowcolor[HTML]{eff0f1}Full mark &-- &-- & 100 & 100 & 10 & 100 &100 &100 &10 &100 &100 & 100 &10 \\
\rowcolor[HTML]{eff0f1}Random &-- &-- & 25 & 25 & -- & 25 &25 &25 &-- &25 &25 & 25 &-- \\

\hline
 \rowcolor[HTML]{E3F8F8}\multicolumn{14}{l}{\gray{\textit{\textbf{Image MLLMs}}}}\\
 \rowcolor[HTML]{E3F8F8}  Otter-I~\cite{otter2023} &2023-05&16 frm  & 25.0 & 25.0 & 2.18 & 25.1 & 25.0 & 24.9  & 4.12  & 13.1 &25.2 & 23.3 &3.15 \\
          \rowcolor[HTML]{E3F8F8} LLaVA-1.6~\cite{llava2023} &2024-01&16 frm  & 60.6 & 41.0 & 2.11 & 43.1 & 38.4 & 41.0  & 4.35 & 25.5 &25.7 &39.3 &3.23\\
         \rowcolor[HTML]{E3F8F8} Claude-3-Opus$^\dag$~\cite{Claude3} &2024-03&16 frm  &67.2 & 43.5  & 3.11 & 21.6 & 40.2 & 47.8  & 3.66 &18.2 &16.7 &36.5 &3.39  \\
        \rowcolor[HTML]{E3F8F8} Qwen-VL-Max$^\dag$~\cite{qwenvl-2023}&2024-01&16 frm  & 67.4 & 63.5 & 2.71 & 40.3 & 40.9 & 43.3  & 5.21 & 25.0 &14.8 &42.2 &3.96  \\
        \hline
         \rowcolor[HTML]{FFF5F5}\multicolumn{14}{l}{\gray{\textit{\textbf{Short Video MLLMs}}}} \\
        \rowcolor[HTML]{FFF5F5} Otter-V~\cite{otter2023}&2023-05&16 frm & 24.6 & 26.0 & 2.38 & 28.2 & 27.6 & 22.3  & 4.23 & 15.1 &26.7 &24.4 &3.31  \\
        \rowcolor[HTML]{FFF5F5} mPLUG-Owl-V~\cite{mplug-owl-2023}&2023-04&16 frm & 28.0 & 25.0 & 2.36 & 24.5 & 31.8 & 27.3  & 5.31 & 21.2 &23.3 &25.9 &3.84 \\
         \rowcolor[HTML]{FFF5F5} VideoChat~\cite{videochat2023}&2023-05&16 frm  & 33.0& 32.0 & 2.31 & 27.0 & 32.1 & 27.6  & 5.01 & 24.3 &28.6 &29.2 &3.66  \\
        \rowcolor[HTML]{FFF5F5}  Video-LLaMA-2~\cite{videollama}&2024-08&16 frm  & 54.5  &41.5  &2.34  &39.4  &33.5  &35.4   & 5.22  &18.5  &25.7 &35.5 &3.78  \\
         \rowcolor[HTML]{FFF5F5} VideoChat2-HD~\cite{mvbench2023} &2024-06 &16 frm & 74.6 & 51.5 & 2.57 & 42.0 & 47.4 & 43.8  & 5.04 & 22.8 &29.6 &44.5 &3.81  \\
          \rowcolor[HTML]{FFF5F5}  Video-LLaVA~\cite{videollava2023} &2023-11 &8 frm  &71.6  &57.0  &2.43  &53.2 &45.2  &48.4   &5.25  &20.1  &35.9 &47.3 &3.84  \\
          \rowcolor[HTML]{FFF5F5}  ShareGPT4Video~\cite{chen2024sharegpt4video} &2024-05 &16 frm  &75.8  &51.5  &2.52  &47.6 &43.2  &48.4   &5.02  &34.0  &23.3 &46.4 &3.77  \\
\rowcolor[HTML]{FFF5F5}  
VideoLLaMA2~\cite{cheng2024videollama2} &2024-06 &16 frm  &74.6  &64.5  &2.79  &49.9 &43.8  &45.1   &5.18  &34.0  &27.4 &48.5 &3.99  \\
        \hline
        \rowcolor[HTML]{F1F6EC}\multicolumn{14}{l}{\gray{\textit{\textbf{Long Video MLLMs}}}}\\
         \rowcolor[HTML]{F1F6EC}  MovieChat~\cite{moviechat2023} &2023-07 &2048 frm   & 29.5  & 25.0  &2.33  &24.2  &24.7 &25.8  &3.23    &28.6  &22.8 &25.8 &2.78    \\
         \rowcolor[HTML]{F1F6EC}  Movie-LLM~\cite{moviellm2024} &2024-03 &1 fps  & 30.0  & 29.0  &2.88  &29.6  &24.7  &24.1    &5.00  &20.5 &24.8 &26.1 &3.94   \\
           \rowcolor[HTML]{F1F6EC}  TimeChat~\cite{timechat2023} & 2023-12 &96 frm  & 23.1  & 27.0  &2.54  &24.5  &28.4  &25.8   &4.29  &24.7  &32.0 &30.9 &3.42  \\
        \rowcolor[HTML]{F1F6EC}  LLaMA-VID~\cite{llama-vid2023} &2023-11 &1 fps  & 50.8  & 34.5  &3.22  &30.1  &32.7  &32.5    &5.22  &23.9 &27.8 &33.2 &4.22   \\
        \rowcolor[HTML]{F1F6EC}  MA-LMM~\cite{malmm2024} & 2024-04 &1000 frm  & 51.9  & 35.5  &2.12  &43.1  &38.9  &35.8    &4.80  &25.1 &24.3 &36.4 &3.46   \\
          \rowcolor[HTML]{F1F6EC} MiniGPT4-Video~\cite{minigpt4video-2024} & 2024-04 &90 frm  & 70.9  & 52.5  &2.64  &49.0  &48.6  &44.5   &4.07  &23.2 &23.0 &44.5 &3.36   \\
         \rowcolor[HTML]{F1F6EC} LongVA~\cite{zhang2024longva} & 2024-06 &256 frm  & 83.3  & 58.5  &3.39  &69.3  &50.0  &67.2   &5.26  &38.6 &27.2 &56.3 &4.33   \\
         \rowcolor[HTML]{F1F6EC} Video-CCAM~\cite{fei2024videoccam} & 2024-08 &96 frm  & 84.9  &66.0  &2.84  &73.2  &60.5  &66.1   &5.19  &42.1 &38.4 &63.1 &4.01   \\
           \rowcolor[HTML]{F1F6EC} Video-XL~\cite{shu2024videoxl} & 2024-09 &256 frm  & 80.3  & 54.5  &3.25  &73.8  &57.4  &67.9   &5.02  &68.3 &40.3 &64.9 &4.14   \\
        \rowcolor[HTML]{F1F6EC}  GPT-4o$^\dag$~\cite{gpt4o} & 2024-05 &0.5 fps  &{87.4}  &{74.5}  &{4.90}  &{64.8}  &{57.1}  &{65.1}    &{6.69}  &{56.7} &{46.3} &{64.6} &{5.80}  \\
\bottomrule
 \end{tabular}
}
\vspace{5pt}
    \caption{The overall performances on MLVU dev set, including the holistic LVU tasks (TR: Topic Reasoning, AR: Anomaly Recognition, VS: Video Summary), the single-detail LVU tasks (NQA: Needle QA, ER: Ego Reasoning, PQA: Plot QA, SSC: Sub-Scene Captioning), and multi-detail LVU tasks (AO: Action Order, AC: Action Count). M-Avg: the average performance of multiple-choice tasks; G-Avg: the average performance of generation tasks (marked by $*$). Two input strategies are used by the MLLMs in evaluation: Uniform Sampling (\textbf{N frm}), which evenly samples N frames from the video; Frame Rate Sampling (\textbf{N fps}), which samples N frames per second. $\dag$ denotes proprietary models.} 
\label{tab:overall}
\end{table*}

\newpage

\section{Annotation Details of MLVU}
\label{appendix:anno-detail}

\subsection{Topic Reasoning (TR).} 
\begin{figure*}[h]
    \centering
    \includegraphics[width=1\textwidth]{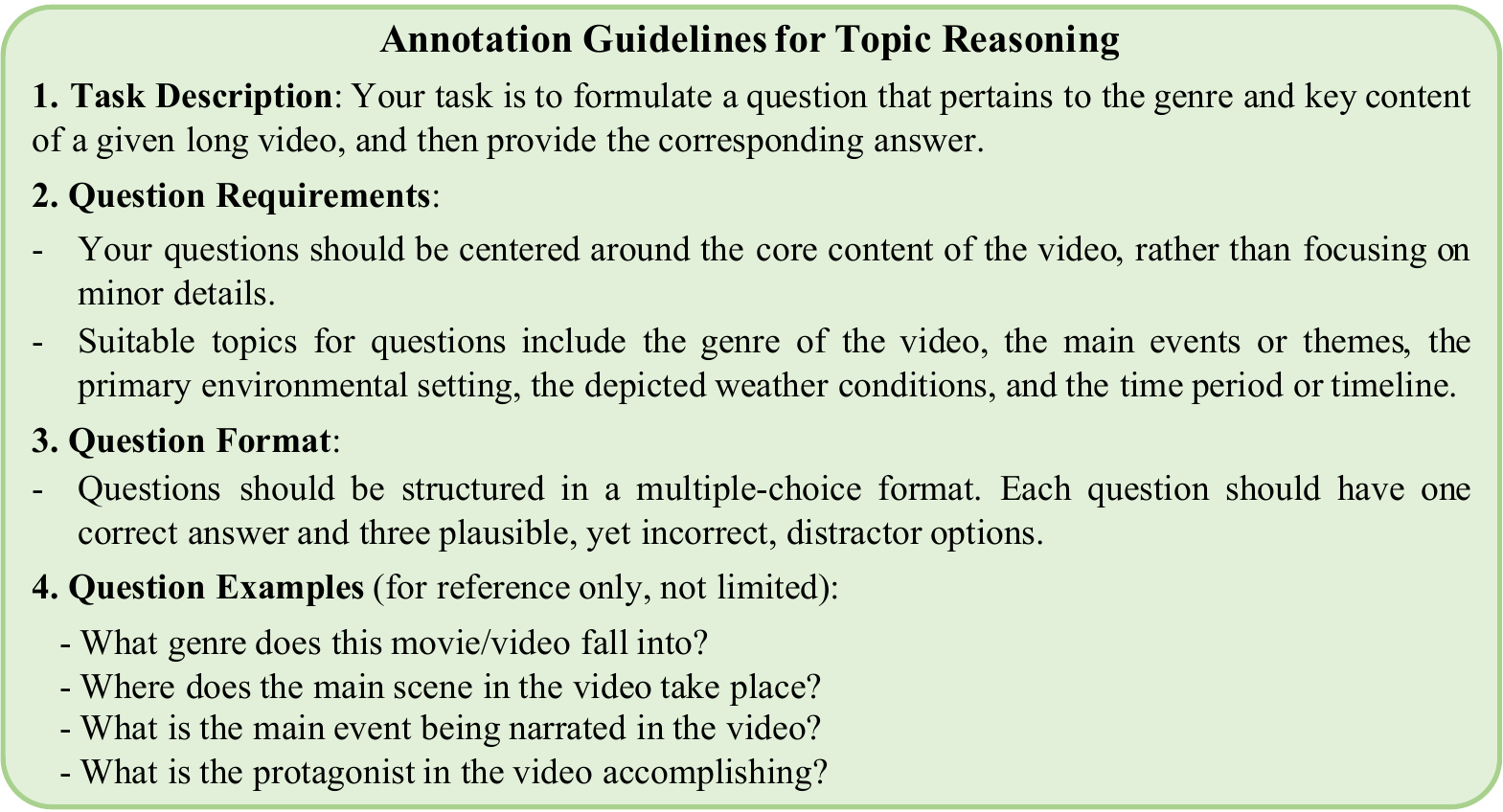}
    \caption{Annotation Guidelines for the Topic Reasoning Task.}
    \label{fig:anno_guide_topic_reason}
\end{figure*}

The questions and corresponding answers for the TR task were meticulously annotated by human annotators, following the specific guidelines illustrated in \Cref{fig:anno_guide_topic_reason}. We required the annotators to design questions related to the reasoning of the video topic, rather than focusing on the creation of questions about minor details. More visualized examples of TR task can be found in \Cref{fig:example-holistic}.

\subsection{Anomaly Recognition (AR).}
The anomaly recognition task did not involve manual annotation. We utilized videos exceeding three minutes in duration, extracted from the UCF-Crime dataset~\cite{ucfcrime-2018}. We also modified the original labels to fit a multiple-choice format.

\subsection{Video Summarization (VS).} 
\label{sub-appendix:anno-detail-VS}
\begin{figure*}[h]
    \centering
    \includegraphics[width=1\textwidth]{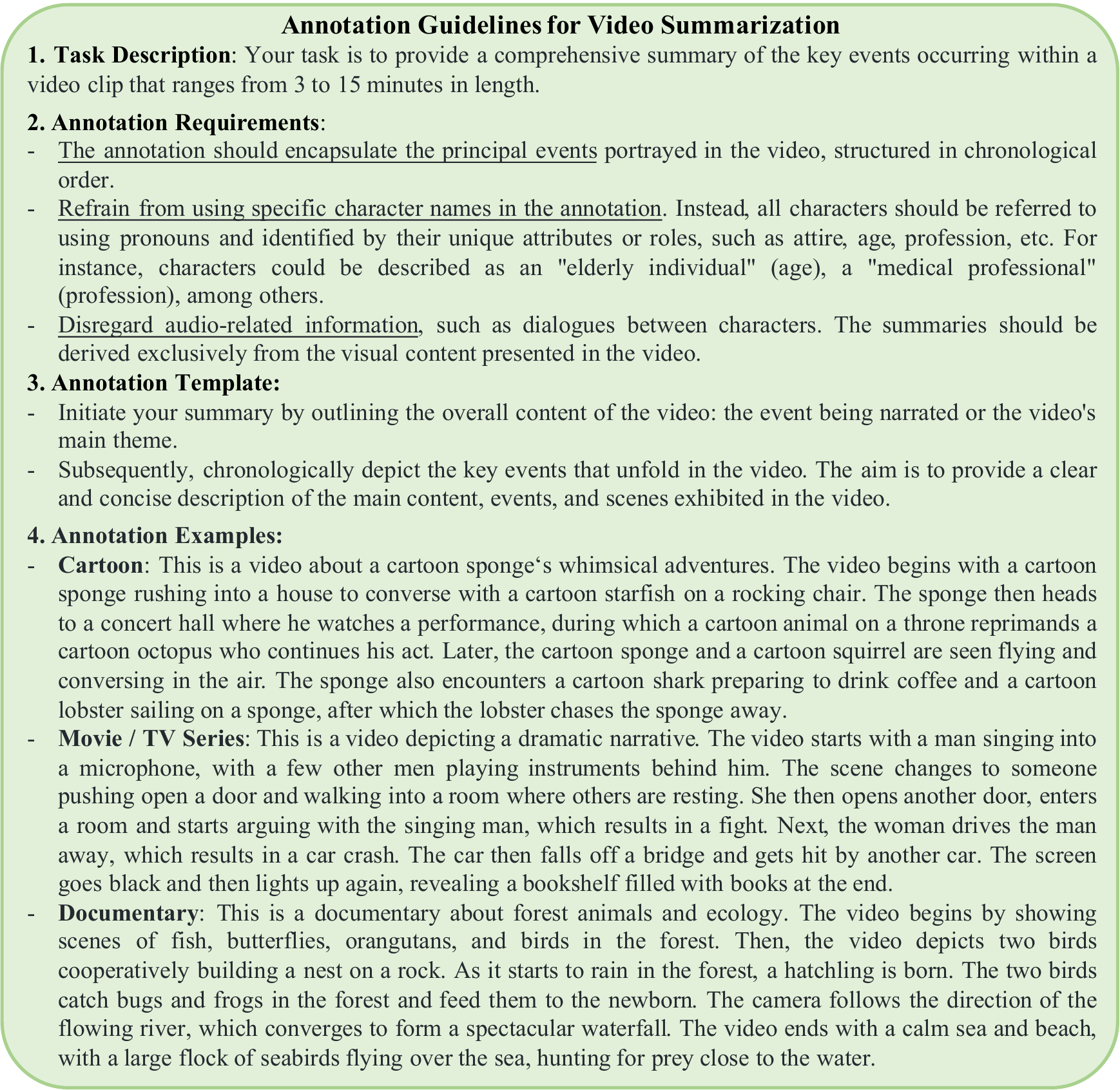}
    \caption{Annotation Guidelines for the Video Summarization Task.}
    \label{fig:anno_guide_video_summary}
\end{figure*}

The ground truth data for the VS task were derived from manual annotations. We instructed the annotators to use pronouns instead of specific character names in all annotations. This guideline stemmed from the inherent constraints of most existing MLLMs, which generally lacked the capacity to process audio or subtitles. This made it difficult for these models to identify specific characters. The annotation instructions and examples provided to the annotators are elaborated in \Cref{fig:anno_guide_video_summary}. More visualized examples of VS task can be found in \Cref{fig:example-holistic}.

\subsection{Needle Question-Answering (NQA).}
\begin{figure}[h]
    \centering
    \includegraphics[width=0.46\textwidth]{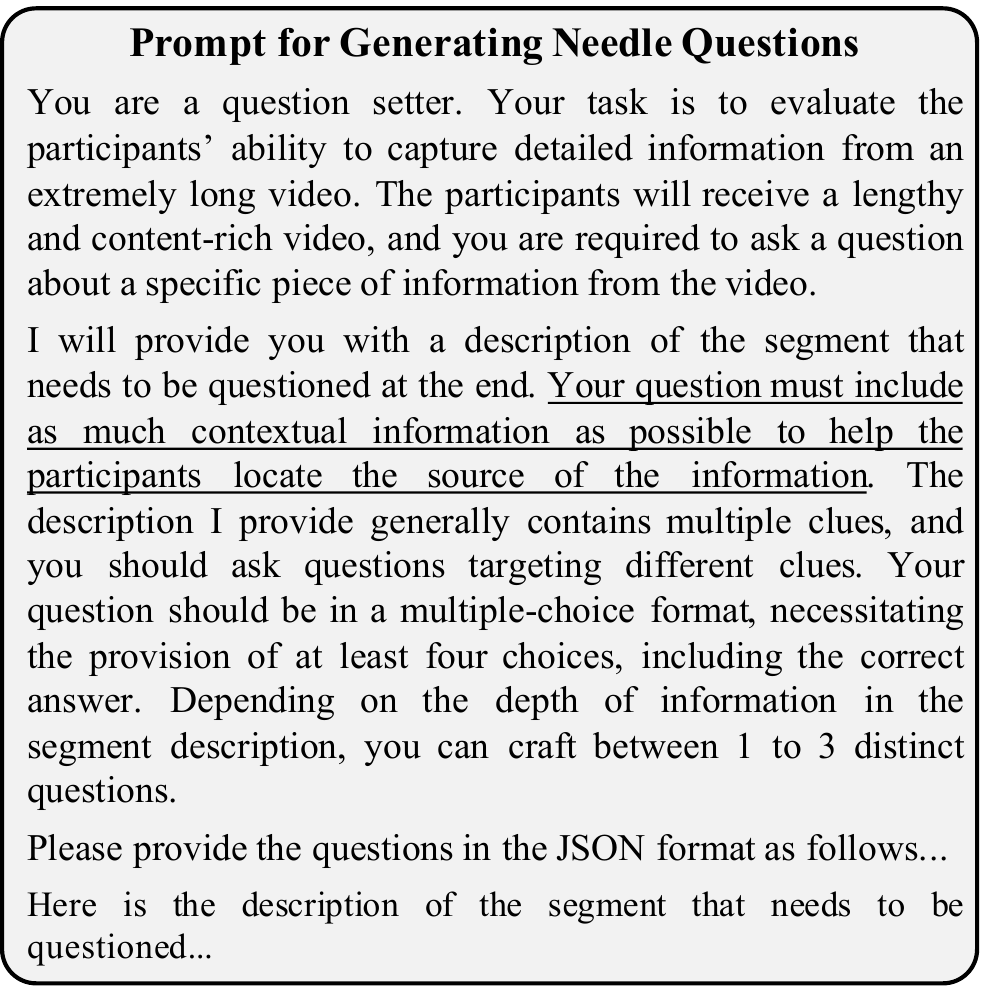}
    \caption{The prompt provided to GPT-4 in the process of creating the question-answer pair for the Needle Question-Answering task.}
    \label{fig:prompt_for_needle_creation}
\end{figure}

We leveraged the GPT-4~\cite{achiam2023gpt} and the detailed video caption data from the WebVid dataset~\cite{webvid-2021} to facilitate a semi-automated generation of annotated questions and answers for the NQA task. Initially, we selected video clips from WebVid, which we refered to as \textit{needle} clips. The corresponding captions of these needle clips were then fed into GPT-4, which generated question-answer pairs based on the information encapsulated in the captions.
The specific prompt provided to GPT-4 is depicted in \Cref{fig:prompt_for_needle_creation}. The generated questions were carefully crafted to focus on a particular detail within the needle clip. These questions were structured to incorporate the maximum number of hints to effectively guide MLLMs in grounding the content of the needle within the context of the longer video. 
Following this, we randomly selected longer background videos from our ULVC and manually ensured that the scene indicated by the needle's question did not feature in these background videos. The final step involves integrating the needle into the longer video, thereby producing the final needle question video. More visualized examples of NQA task can be found in \Cref{fig:example-single-detail}.

\subsection{Ego Reasoning (ER).}
The video resources, questions, and correct responses used in the ER task were derived from the Natural Language Queries (NLQ) task within the Ego4D dataset~\cite{ego4d2022}. This data was restructured to fit a multiple-choice question format.

\subsection{Plot Question-Answering (PQA).}
\label{sub-appendix:anno-detail-PQA}
\begin{figure*}[h]
    \centering
    \includegraphics[width=1\textwidth]{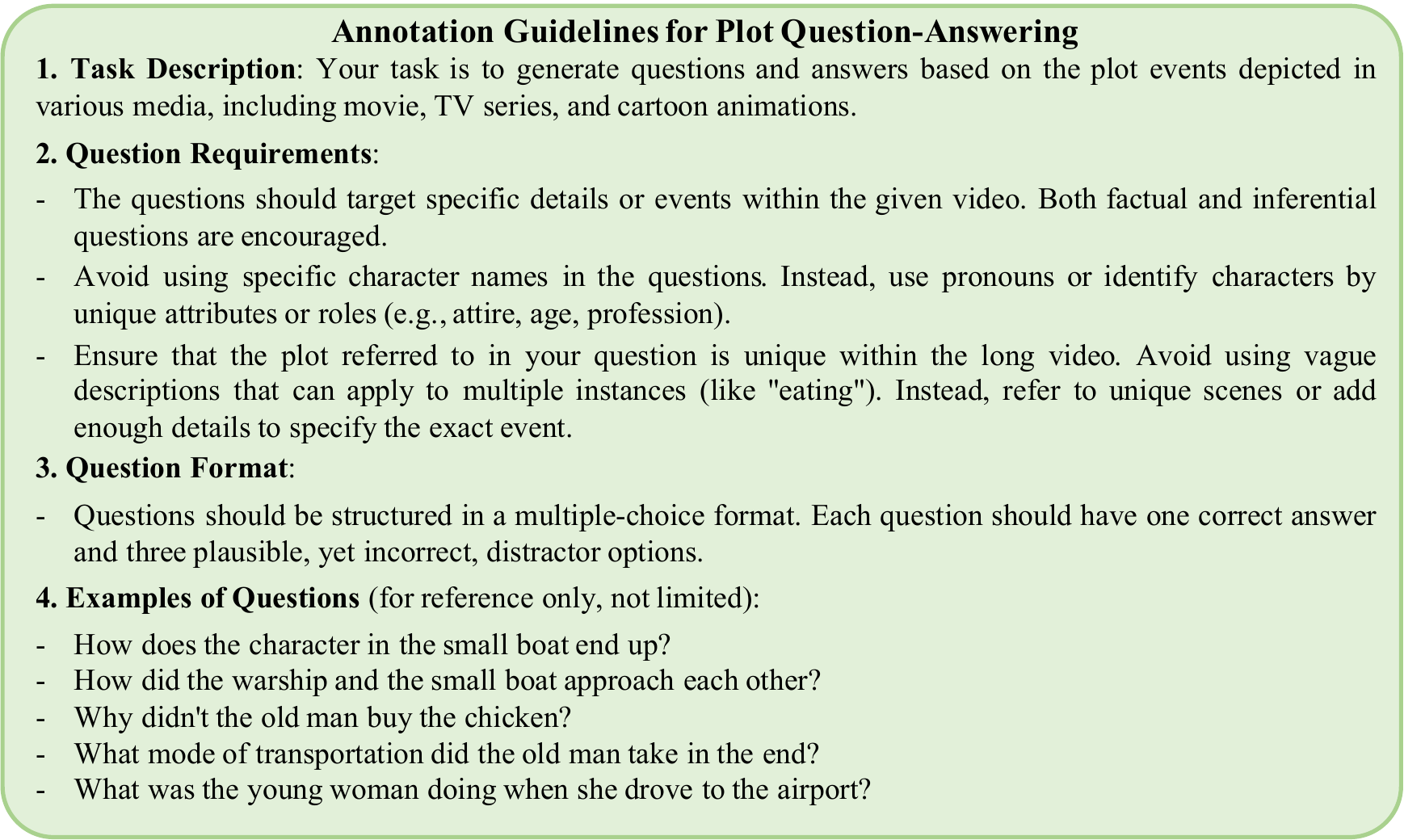}
    \caption{Annotation Guidelines for the Plot Question-Answering Task.}
    \label{fig:anno_guide_PQA}
\end{figure*}

The PQA task's questions and answers were annotated by human annotators, following specific guidelines illustrated in \Cref{fig:anno_guide_PQA}. We instructed the annotators to craft questions that probe into the intricate plot details encapsulated within the videos. These questions were designed to encompass both perception and reasoning aspects. We stipulated that both questions and their corresponding answers should avoid the use of specific character names or any objective hints, and should instead utilize pronouns. This approach was strategized to prevent potential information leakage, given that MLLMs often demonstrate a familiarity with the storylines of well-known movies and TV series. Such common-sense knowledge could potentially allow the MLLMs to answer questions correctly without the essential requirement of analyzing the input video. 

\begin{figure*}[h!]
    \centering
    \includegraphics[scale=0.215]{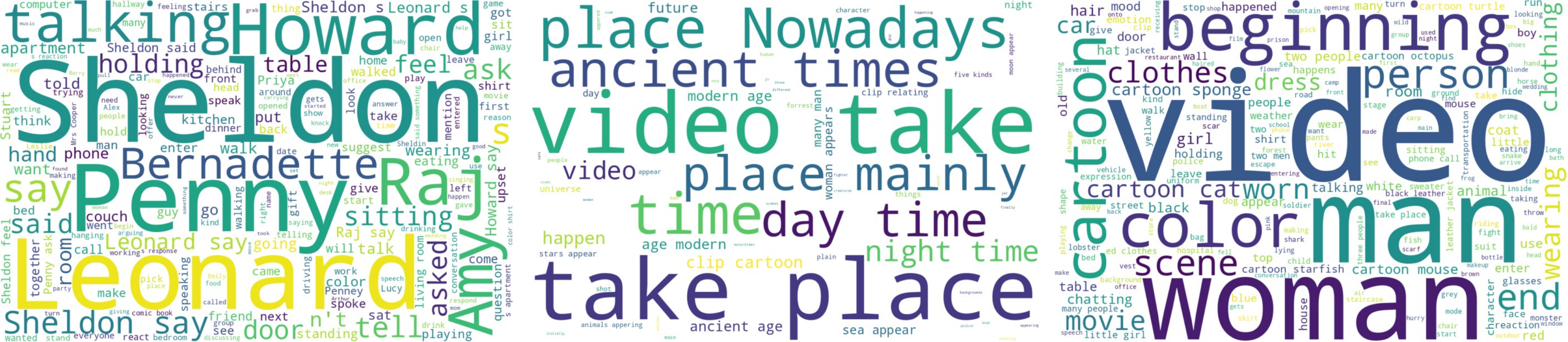}
    \caption{\textbf{Word Cloud Comparison} of questions in TVQA test set (left), MovieChat test set (middle), and our PQA (right). Notably, TVQA's character-specific names require LLMs to recognize characters, risking reliance on pre-existing knowledge. In contrast, MovieChat questions are less diverse. Our PQA addresses these issues, providing enhanced usability and reliability.}
    \label{fig:plotqa-wordcloud}
\end{figure*}

Nonetheless, the complexity of character interactions and actions in longer videos poses a challenge to conveying plot details using only pronouns and feature descriptions.  Previous datasets for plot question answering that avoided the use of character names often resulted in compromised question diversity and tended towards generalized queries. We illustrate this through a comparative analysis of TVQA \cite{tvqa2018}, Moviechat \cite{moviechat2023}, and our PQA dataset's question word clouds in \Cref{fig:plotqa-wordcloud}. While TVQA provides a diverse range of questions, it does so by employing specific character names. In contrast, Moviechat avoids character names, but its questions are frequently overly broad, lack specific plot details, and exhibit diminished diversity. Our PQA dataset successfully navigates these challenges, offering a diverse range of questions without resorting to the use of character names. More visualized examples of PQA task can be found in \Cref{fig:example-single-detail}.


\subsection{Sub-Scene Captioning (SSC).}
\label{sub-appendix:anno-detail-SSC}
\begin{figure*}[h]
    \centering
    \includegraphics[width=1\textwidth]{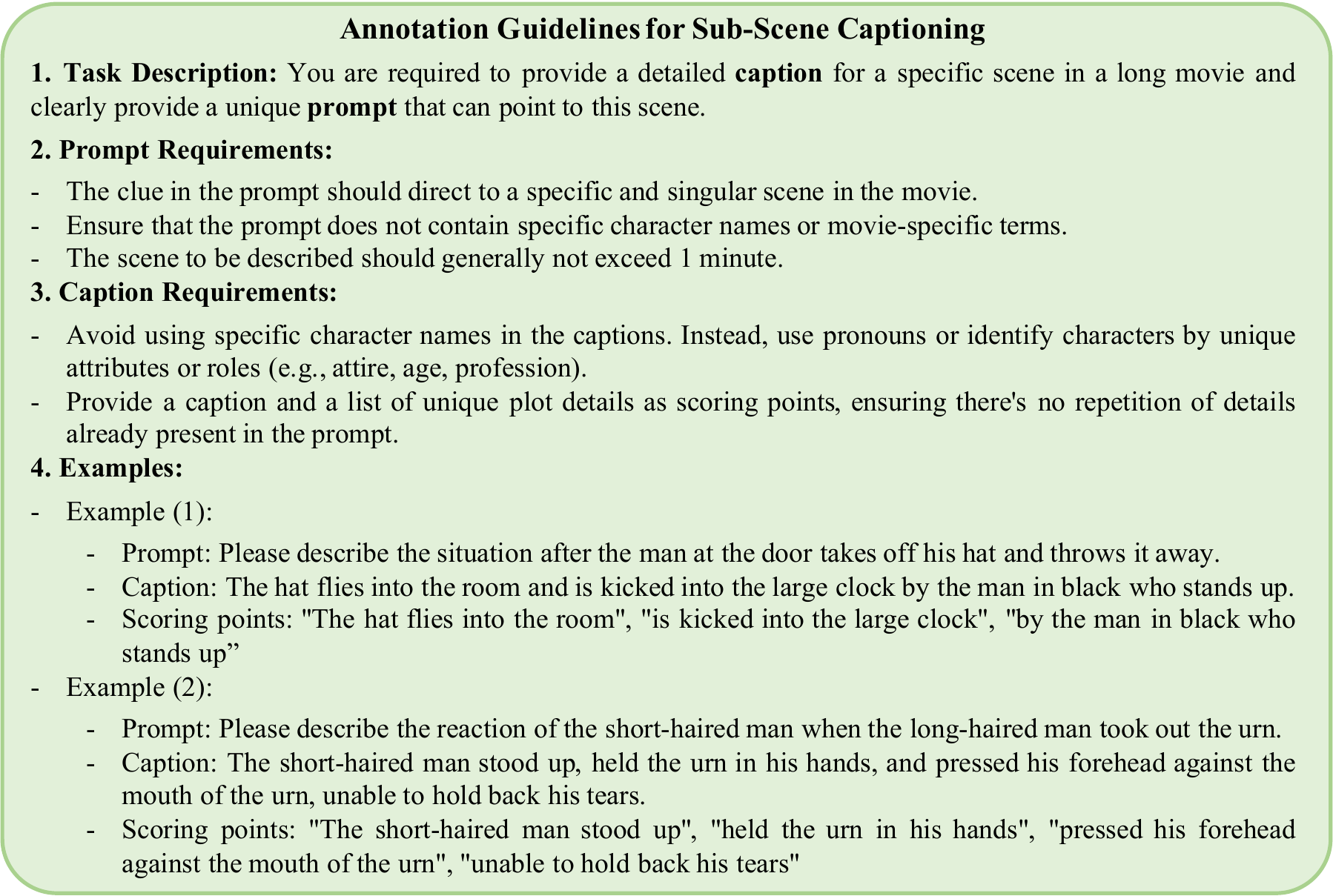}
    \caption{Annotation Guidelines for the Sub-Scene Captioning Task.}
    \label{fig:anno_guide_SSC}
\end{figure*}

In the development process of the SSC task, we employed human annotators to generate both prompts and standard caption data. The specific guidelines provided to annotators are illustrated in \Cref{fig:anno_guide_SSC}. Initially, the annotators identified a specific, easily referable sub-scene within a lengthy movie. Subsequently, they crafted a prompt replete with adequate clues to reference this scene, ensuring the uniqueness of these clues throughout the entire film. To prevent any leakage of information, the prompt was designed to exclude any character-specific names or objective hints, instead incorporating rich descriptive details to allude to the plot. Following this, the annotators produced a detailed caption for this sub-scene, and deconstructed the caption into multiple, non-redundant "scoring points" to facilitate quantitative assessment (the details of the evaluation metric can be found in \Cref{sub-appendix:evaluation_metric}). More visualized examples of PQA task can be found in \Cref{fig:example-ssc-detail}.

\subsection{Action Order (AO).} 
The videos, questions, and answers for the action order task were all synthetically generated. In order to maintain the high quality of our evaluation data, we adopted a dual-strategy approach. Firstly, we selected actions for the \textit{probe} videos that were not commonly seen in most films, such as making jewelry and water skiing. Secondly, in the selection of background videos, we conducted a cursory review of the video content to further ensure that the actions referenced in the questions were not present in the video. This rigorous methodology ensured the reliability of our data.

\subsection{Action Count (AC).} 
The process of data acquisition and annotation for the action count task closely mirrored that of the action order task. All videos, questions, and answers were synthetically generated. We employed a strategy consistent with the action order task to ensure the validity and reliability of our evaluation data.

\section{Details of Baselines and the Evaluation Process}
\label{appendix:evaluation}
\subsection{Baselines}
\label{sub-appendix:evaluation-baselines}
In this section, we outline the primary baselines evaluated on our MLVU. For image-based MLLMs, most models lack multi-image inference capabilities. Therefore, we select Otter-I, LLaVA-1.6, and InternVL, which have official multi-image implementations. Additionally, we include two proprietary models—Claude-3-Opus and Qwen-VL-Max—that offer APIs for multi-image inference. For the available models, we determine the maximum input frames based on their LLM context length. Claude and Qwen support a maximum of approximately 20 images, so we choose 16 frames to ensure fair comparisons. Regarding video MLLMs, we use default frame sampling strategies. For example, VideoChat2 uniformly samples 16 frames, while LLaMA-Vid samples 1 frame per second. Specifically, GPT-4o can handle up to approximately 500 images at a resolution of 512×512 pixels. Thus, we select a sampling rate of 0.5 fps to accommodate most of our videos.

\subsection{Inference Detatils} 
\begin{figure*}[h]
    \centering
    \includegraphics[width=0.8\textwidth]{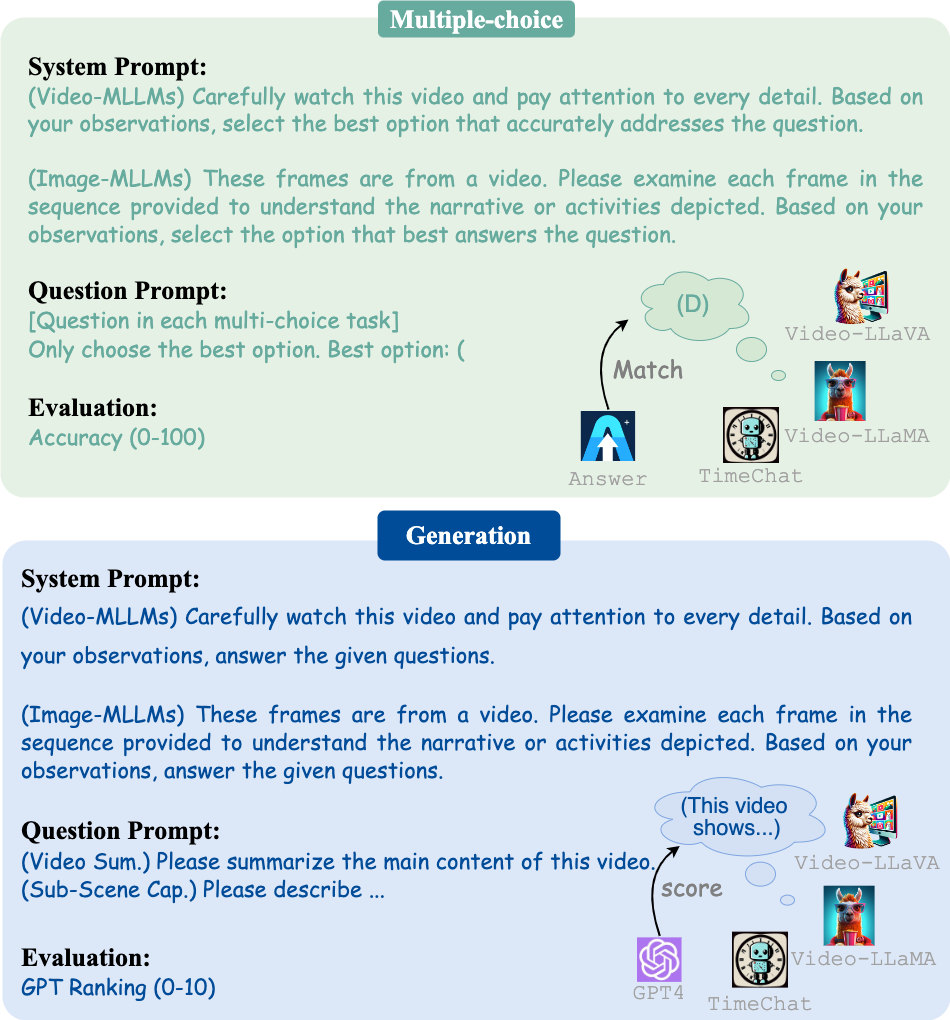}
    \caption{Inference template for our MLVU. }
    \label{fig:infer}
    \vspace{-0.3cm}
\end{figure*}

We have developed two templates specifically for Multiple-Choice and Generation tasks, as illustrated in Figure \ref{fig:infer}. Distinct system prompts were designed to accommodate the differences between video-based and image-based MLLMs. Considering the variances in task requirements, we incorporated ``option prediction guidance'' into the Multiple-Choice template to aid in option extraction. Conversely, in Generation tasks, we do not implement any additional interventions but employ fixed-question guidance to enable models to respond to diverse task questions. In our evaluation, the templates are seamlessly integrated into the evaluation code of open-release models or available API of proprietary models.

\subsection{Evaluation Metrics} 
\label{sub-appendix:evaluation_metric}
\begin{figure*}[h]
    \centering
    \includegraphics[width=1\textwidth]{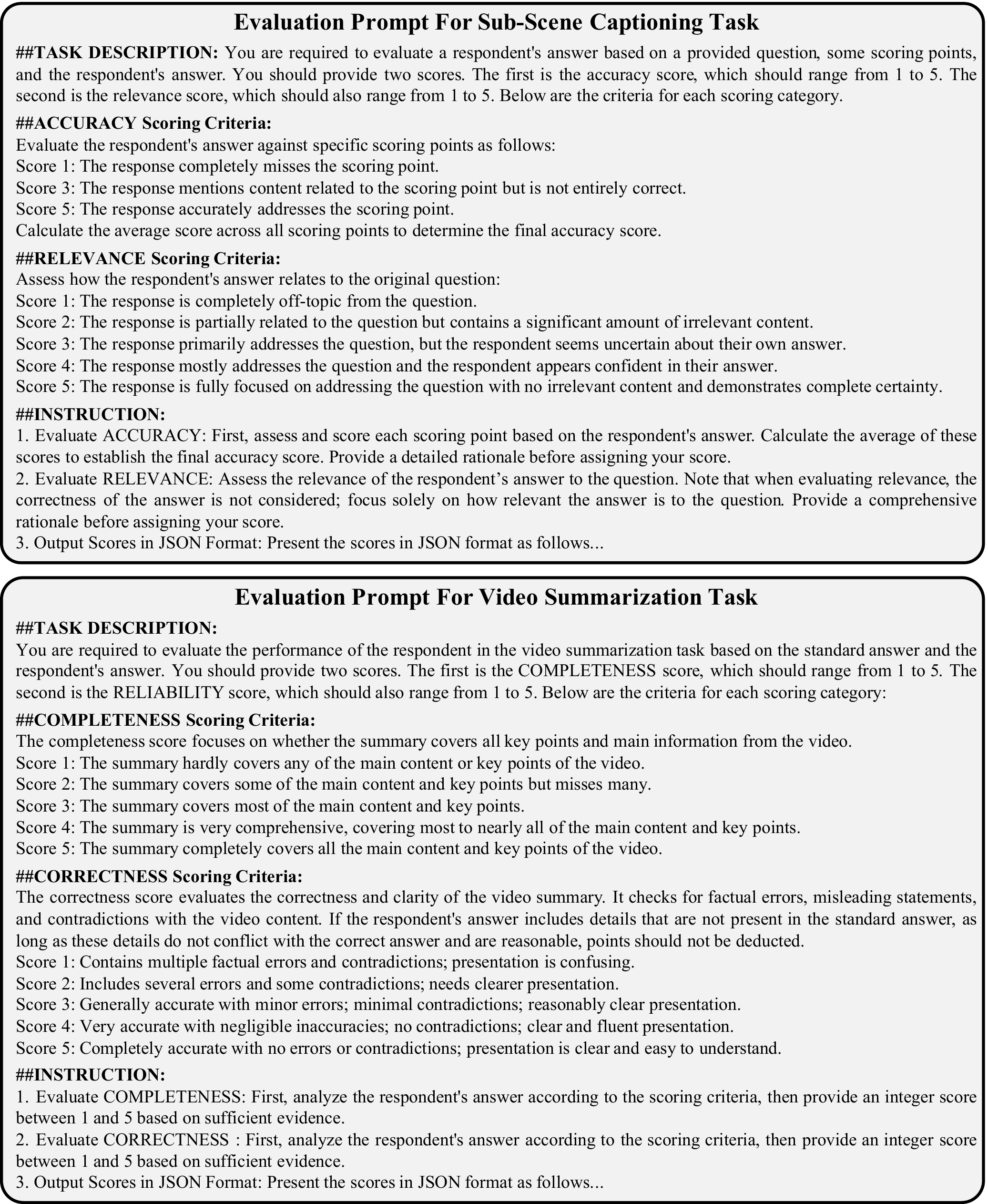}
    \caption{Detailed prompt for evaluation of generation tasks in MLVU. }
    \label{fig:prompt}
    \vspace{-0.3cm}
\end{figure*}

For the evaluation of Multiple Choice tasks, we directly compute absolute accuracy by matching the predicted option with the ground truth. In Generation tasks, we develop multiple criteria for assessment and employ GPT-4 to rank the alignment between generated texts and the provided answers. As illustrated in Figure \ref{fig:prompt}, we use ``Accuracy'' and ``Relevance'' to benchmark Sub-scene Captioning, and ``Completeness'' and ``Reliability'' to evaluate the capabilities of Video Summary.

\section{Explorations of Video Retrieval Augmented Generation}
\label{appendix:rag}

As discussed in Section \ref{sec-exp-detail-analysis}, most MLLMs are adversely affected by video length. Drawing inspiration from the use of Retrieval Augmented Generation (RAG) in video understanding, we have developed a zero-shot RAG strategy and seamlessly integrated it into existing MLLMs. Table \ref{tab:rag_full} displays the performance comparison between the baseline models and the models employing our RAG strategy. It is noteworthy that all methods benefit from the RAG strategy in Needle QA, Ego Reasoning, and Plot QA. Conversely, minimal improvement is observed in Action Count, and a decrease is noted in Action Order and Overall Reasoning. This is primarily because RAG facilitates the retrieval of detail-oriented video clips, which makes models more likely to focus on answer-related cues in specific single-detail reasoning tasks. However, RAG exhibits limited capabilities in multi-detail reasoning and holistic understanding tasks, which require global perception and knowledge aggregation.

\begin{figure*}[h]
    \centering
    \includegraphics[width=1\textwidth]{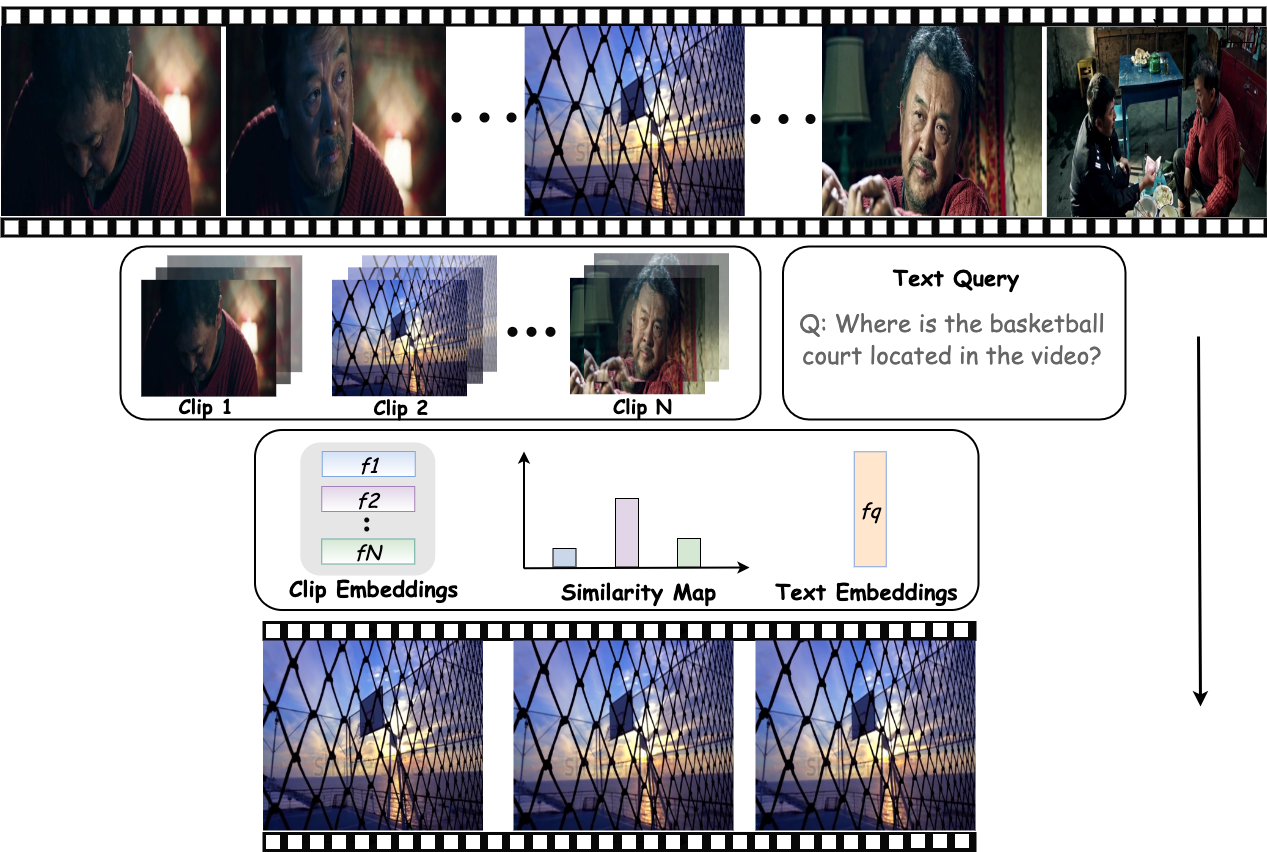}
    \caption{Pipeline of our video retrieval augmented generation strategy. }
    \label{fig:rag}
    \vspace{-0.3cm}
\end{figure*}

The pipeline of our video retrieval augmented generation is illustrated in Figure \ref{fig:rag}. Initially, a long video is uniformly divided into $N$ video clips, each containing $C$ frames. Subsequently, we employ a robust video feature extraction tool, LanguageBind \cite{zhu2023languagebind} to extract clip embeddings $F_I \in \mathbb{R}^{N \times d}$, where $d$ represents the dimension of each clip embedding. We then compute the similarities between $F_I$ and the text embedding $F_T$, concatenating the top $K$ clips to enhance the model's capability for question-answering. Given that many Video MLLMs are limited to processing only 16 frames, we have adjusted the settings for $C$ and $K$ to accommodate video retrieval in 16-second intervals. As discussed below, the RAG strategy excels in detail-oriented tasks but shows limitations in global understanding tasks. Moreover, it is relatively inefficient, requiring more than one minute to complete the process. Consequently, more effective approaches need to be developed for long video understanding tasks, and we aim to address this in future work.

\begin{table*}[h]
    \centering
    \setlength\tabcolsep{2.5pt}
    \resizebox{1.0\linewidth}{!}{
        \begin{tabular}{l c c c c c c c c}
        \Xhline{1.0pt}
       \textbf{Model} & \textbf{Settings} & \textbf{Needle QA} & \textbf{Ego Rea.} & \textbf{Plot QA} & \textbf{Action Or.} & \textbf{Action Co.} & \textbf{Anomaly Rec.}& \textbf{Topic Rea.} \\
        \Xhline{0.8pt}
       \rowcolor{gray!20} LLaVA-B & - & 43.1 & 38.4  & 41.0  & 25.5  & 25.7  & 41.0  & 60.6 \\
       \Xhline{0.8pt}
        \multirow{3}{*}{LLaVA-R} & C=2,K=8 & 50.7 & 45.7 &49.7 &26.3 &26.7 &40.8 &59.8  \\
         & C=4,K=4 & 53.5 & 43.5 & 50.6 & 25.9 &29.6 &39.9 &58.5  \\
         & C=8,K=2 & 55.2 & 42.6 & 50.3 & 25.1 & 30.1 &40.6 &59.5 \\
        \Xhline{0.8pt}
        \rowcolor{gray!20}  InternVL-B & - & 52.7 & 43.5  & 54.4  & 32.8  & 23.8  & 67.0  & 78.8 \\
         \Xhline{0.8pt}
          \multirow{3}{*}{InternVL-R} & C=2,K=8 & 77.2 & 52.6 &61.4 &30.1 &36.4 &57.9 &69.2  \\
         & C=4,K=4 & 76.3 & 51.4 & 59.9 & 29.3 &36.9 &58.3 &69.4  \\
         & C=8,K=2 & 77.8 & 48.9 & 61.6 & 31.7 & 33.0 &60.2 &62.3 \\
        \Xhline{0.8pt}
        \rowcolor{gray!20}  Video-LLaMA-B & - & 39.4 & 33.5  & 35.4  & 18.5  & 25.7  & 41.5 & 54.5 \\
         \Xhline{0.8pt}
           \multirow{3}{*}{Video-LLaMA-R} & C=2,K=8 & 61.4 & 42.6 &38.8 &17.4 &17.5 &35.7 &48.5  \\
         & C=4,K=4 & 58.9 & 42.6 & 39.1 & 17.8 &23.8 &36.0 &49.3  \\
         & C=8,K=2 & 62.0 & 38.4 & 36.2 & 25.5 & 18.0 &38.5 &51.0 \\
        \Xhline{0.8pt}
        \rowcolor{gray!20}  VideoChat2-B & - & 42.0 & 47.4  & 43.8  & 22.8  & 29.6  & 51.5  & 74.6 \\
         \Xhline{0.8pt}
         \multirow{3}{*}{VideoChat2-R} & C=2,K=8 & 72.1 & 53.7 &55.5 &21.6 &30.1 &45.8 &68.2  \\
         & C=4,K=4 & 72.4 & 55.4 & 53.4 & 22.4 &31.1 &45.3 &68.9  \\
         & C=8,K=2 & 73.8 & 53.1 & 55.3 & 22.0 &31.6 &46.6 &69.7 \\
        \Xhline{0.8pt}
        \rowcolor{gray!20}  MiniGPT4-Video-B & - & 49.0 & 48.6  & 44.5  & 23.2  & 23.0  & 52.5  & 70.9 \\
         \Xhline{0.8pt}
           \multirow{3}{*}{MiniGPT4-Video-R} & C=2,K=8 &60.6  &44.3  &47.4 &23.2 &23.7 &42.8 &60.9  \\
         & C=4,K=4 &60.3  &44.6  &46.9  &26.3  &23.8 &42.6 &60.7  \\
         & C=8,K=2 &56.3  &44.6  &46.6  &27.4  &24.8 &45.0 &47.5 \\
        \Xhline{1.0pt}
        \end{tabular}
    }
    \caption{Quantitative results on video Retrieval Augmented Generation. ``model-B'' and ``model-R'' denote Baseline and RAG models respectively. We evaluate two image MLLMs and three video MLLMs in different settings.}  
    \label{tab:rag_full}
    \vspace{-0.4cm}
\end{table*}

\section{More Visualized Examples of MLVU.}
\label{appendix:more-mlvu-example}

\begin{figure*}[h]
    \centering
    \includegraphics[width=0.86\textwidth]{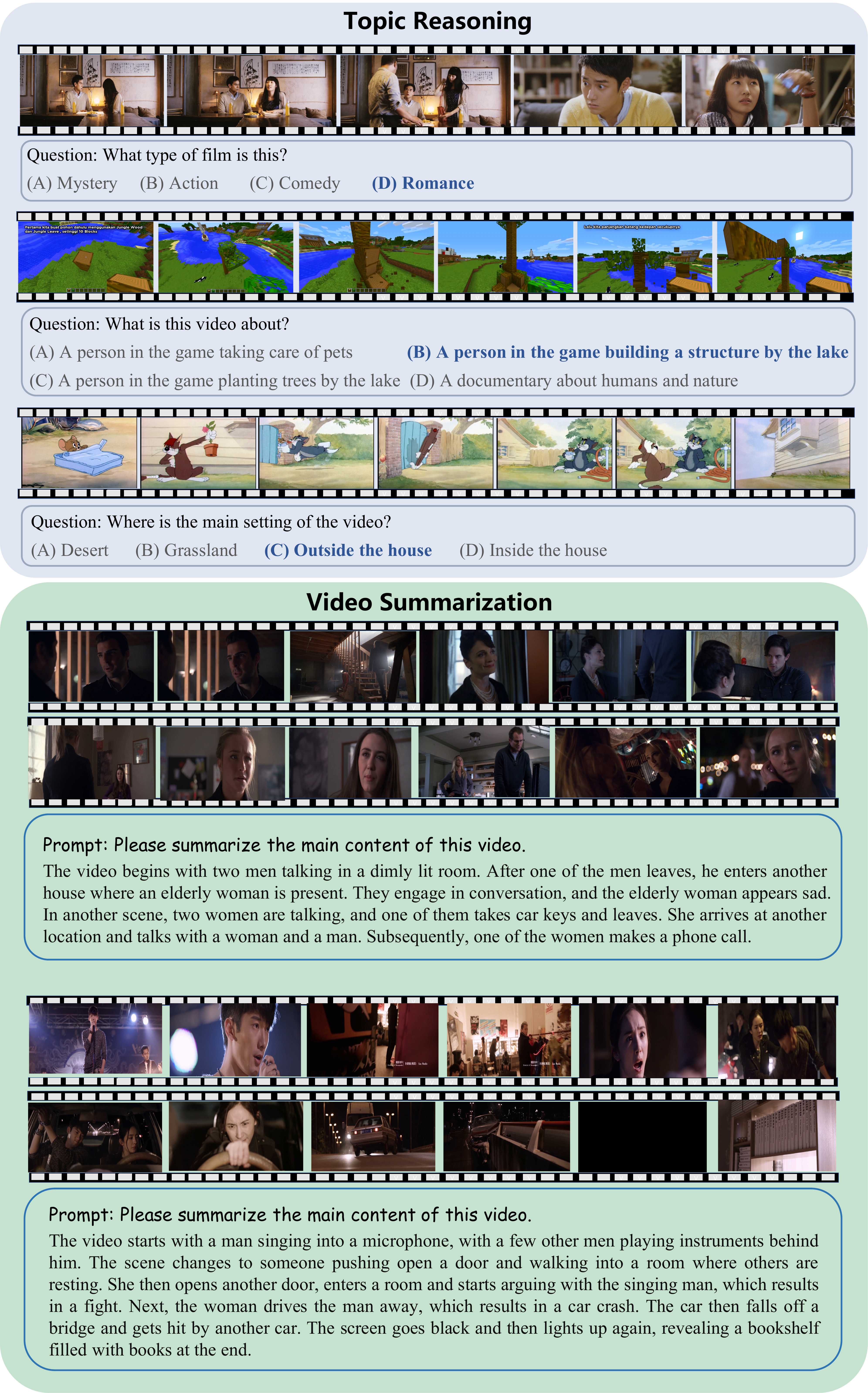}
    \caption{More Examples of Topic Reasoning and Video Summarization Tasks.} 
    \label{fig:example-holistic}
\end{figure*}

\begin{figure*}[h]
    \centering
    \includegraphics[width=0.86\textwidth]{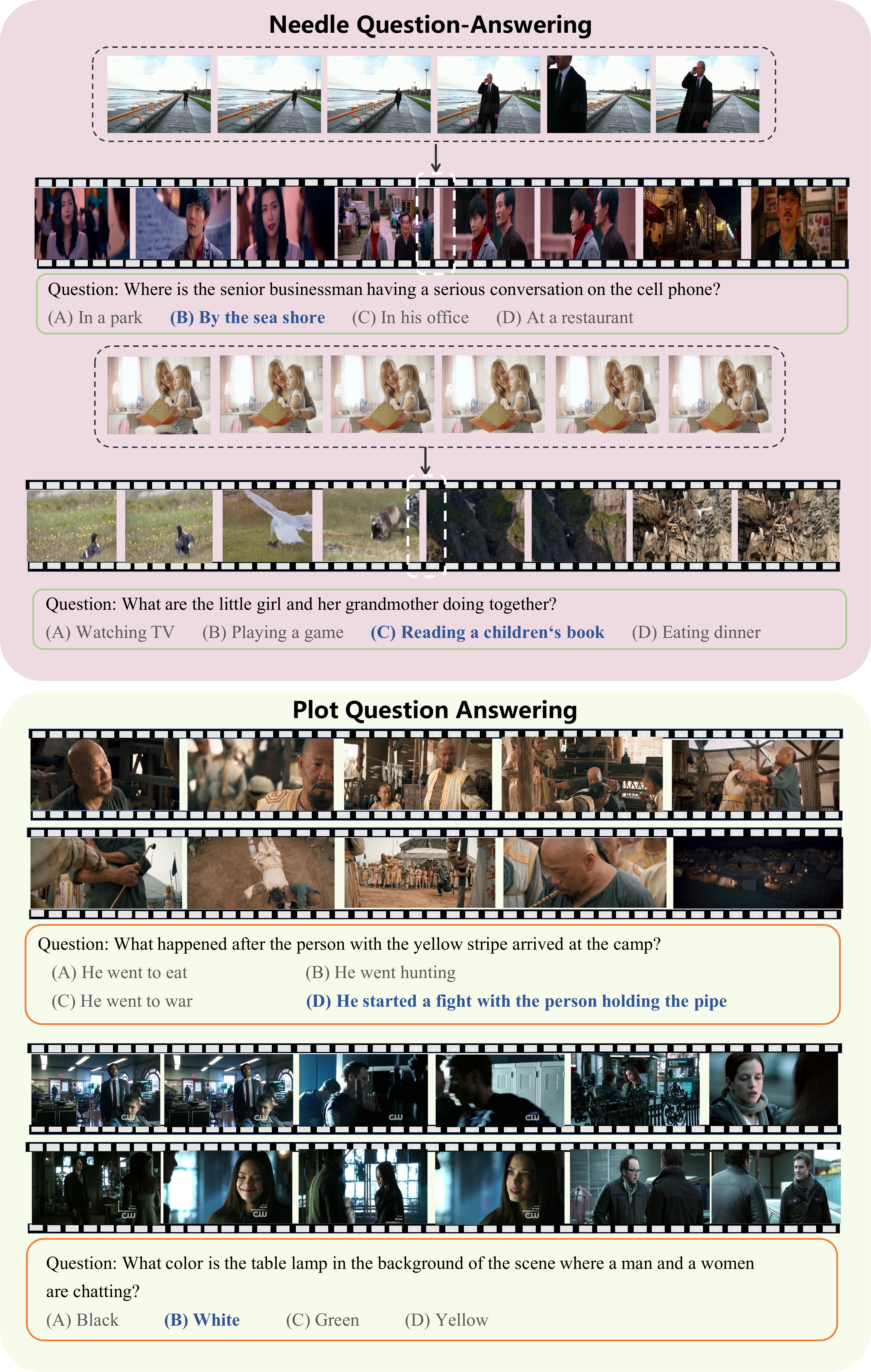}
    \caption{More Examples of Needle Question Answering and Plot Question Answering Tasks.} 
    \label{fig:example-single-detail}
\end{figure*}

\begin{figure*}[h]
    \centering
    \includegraphics[width=0.86\textwidth]{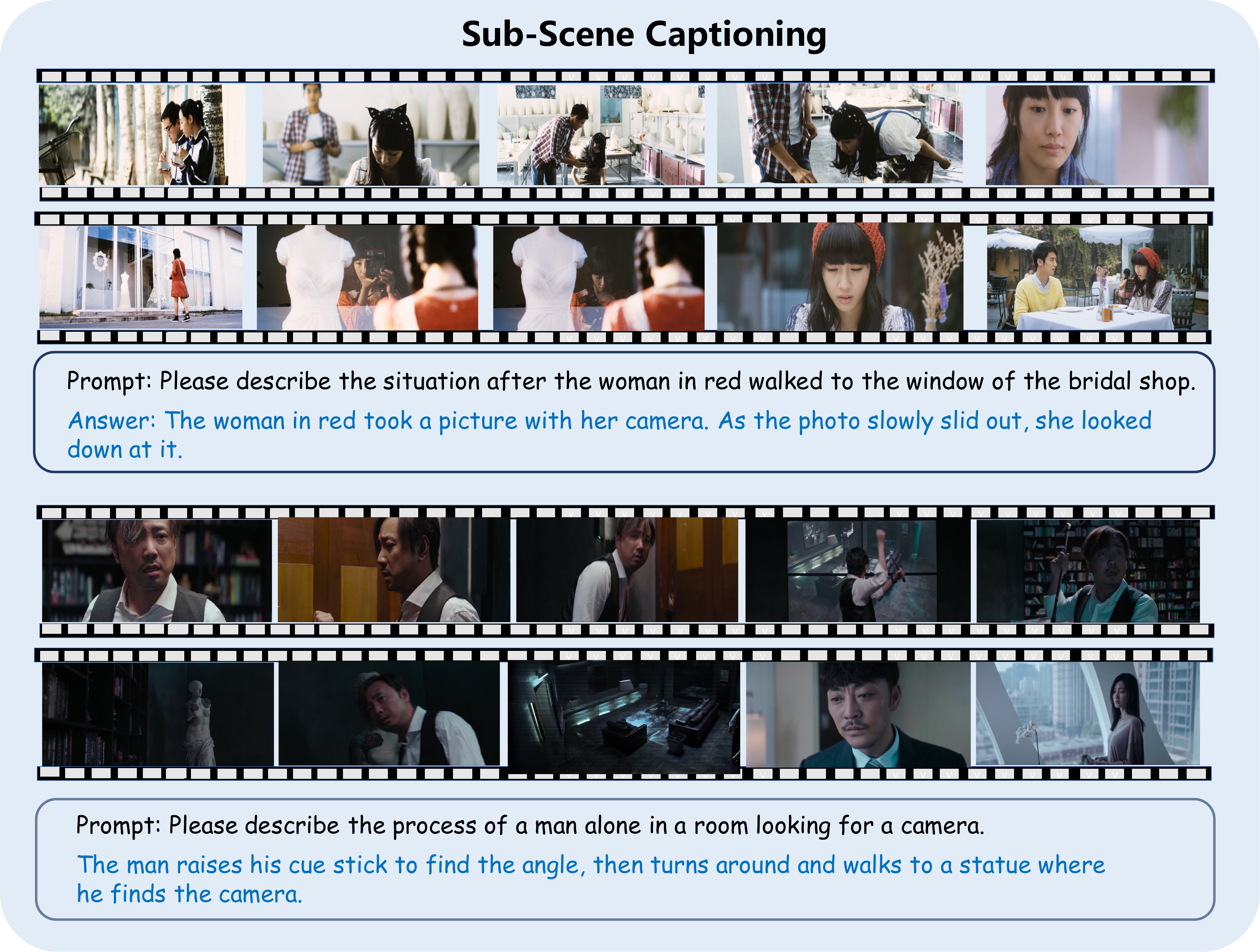}
    \caption{More Examples of Sub-Scene Captioning.} 
    \label{fig:example-ssc-detail}
\end{figure*}

We present additional visualizations of our MLVU annotation examples in Figures~\ref{fig:example-holistic}, \ref{fig:example-single-detail}, and \ref{fig:example-ssc-detail}.

\end{document}